\documentclass[journal]{IEEEtran}
\ifCLASSINFOpdf
\else
\fi
\usepackage{graphicx}
\usepackage{times}
\usepackage{epsfig}
\usepackage{graphicx}
\usepackage{amsmath}
\usepackage{amssymb}
\usepackage{soul}
\usepackage{rotating}
\usepackage{commath}
\usepackage{textcomp}
\usepackage[linesnumbered,ruled]{algorithm2e}
\usepackage{multirow}
\usepackage{placeins}
\usepackage{caption}
\usepackage{subcaption}
\usepackage{mathrsfs}
\usepackage{amsfonts}
\usepackage[table,xcdraw]{xcolor}
\usepackage{lipsum}
\usepackage{widetext}
\usepackage[justification=centering]{caption}

\usepackage[breaklinks=true,bookmarks=false]{hyperref}
\usepackage{cleveref}

\usepackage{mathtools}

\DeclarePairedDelimiter\floor{\lfloor}{\rfloor}

\begin{document}
% Do not put math or special symbols in the title.A CNN With Multiscale Convolution and Diversified Metric for Hyperspectral Image Classification
\title{A CNN With Multiscale Convolution for Hyperspectral Image Classification using Target-Pixel-Orientation scheme}


\author{\IEEEauthorblockN{Jayasree Saha\IEEEauthorrefmark{1},
Yuvraj Khanna\IEEEauthorrefmark{1}, and
Jayanta Mukherjee\IEEEauthorrefmark{1}
}
\IEEEauthorblockA{\\\IEEEauthorrefmark{1} Computer Science and Engineering,
Indian Institute of Technology, Kharagpur, West Bengal, India}
%\IEEEauthorblockA{\IEEEauthorrefmark{1}Twentieth Century Fox, Springfield, USA}
%\IEEEauthorblockA{\IEEEauthorrefmark{1}Starfleet Academy, San Francisco, CA 96678 USA}
%\IEEEauthorblockA{\IEEEauthorrefmark{1}}% <-this % stops an unwanted space
%\thanks{Manuscript received December 1, 2012; revised August 26, 2015. 
%Corresponding author: M. Shell (email: http://www.michaelshell.org/contact.html).}
}

%\author{\IEEEauthorblockN{Jayasree Saha,
%		Yuvraj Khanna,
%		 and
%		Jayanta Mukherjee}
%	\IEEEauthorblockA{\IEEEauthorrefmark{1}\\Department of Computer science and Engineering,
%		Indian Institute of Technology, Kharagpur, India}
%
%
%% <-this % stops an unwanted space
%}

% The paper headers................****************************...........................
%{Shell \MakeLowercase{\textit{et al.}}: Bare Demo of IEEEtran.cls for IEEE Transactions on Magnetics Journals}



% for Transactions on Magnetics papers, we must declare the abstract and
% index terms PRIOR to the title within the \IEEEtitleabstractindextext
% IEEEtran command as these need to go into the title area created by
\IEEEtitleabstractindextext{%
\begin{abstract}
	
Presently, CNN is a popular choice to handle the hyperspectral image classification challenges. In spite of having such large spectral information in Hyper-Spectral Image(s) (HSI), it creates a curse of dimensionality. Also, large spatial variability of spectral signature adds more difficulty in classification problem.
Additionally, scarcity of training examples is a bottleneck in using a CNN. In this paper, a novel target-patch-orientation method is proposed to train a CNN based network. Also, we have introduced a hybrid of 3D-CNN and 2D-CNN based network architecture to implement band reduction and feature extraction methods, respectively.  %The proposed architecture is trained in an end to end manner to accomplish both band reduction as well as feature extraction task.  
Experimental results show that our method outperforms the accuracies reported in the existing state of the art methods.
\end{abstract}

}

\maketitle

\IEEEdisplaynontitleabstractindextext

% \fi
\IEEEpeerreviewmaketitle

\section{Introduction}
\IEEEPARstart{H}{yperspectral} image (HSI) classification has received considerable attention in recent years for a variety of application using neural network-based techniques. Hyperspectral imagery has several hundreds of contiguous narrow spectral bands from the visible to the infrared frequency in the entire electromagnetic spectrum. %It is expected to provide finer classification with such high spectral resolution due to having a distinct spectral signature for each pixel instances. 
However, processing such a large number of spectral dimension suffers from the curse of dimensionality. Also, a few properties of the dataset bring challenges in the classification of HSIs, such as 1) limited training examples, and
2) large spatial variability of spectral signature. In general, contiguous spectral bands may contain some redundant information which leads to the Hughes phenomenon~\cite{Hughes}. It causes accuracy drop in classification when there is an imbalance between the high number of spectral channels and scarce training examples. Conventionally, dimension-reduction techniques are used to extract a suitable spectral feature. For instance, Independent Component Discriminant Analysis (ICDA)~\cite{dimred_villa11_ica} has been used  to find statistically independent components. It uses ICA with the assumption that at most one component has a Gaussian distribution. ICA uses higher order  statistics to compute uncorrelated components compared to the PCA~\cite{dimred_Licciardi_pca} which uses the covariance matrix. 
A few other non-linear techniques quadratic discriminant  analysis~\cite{dimred_Li_quadratic}, kernel-based methods~\cite{dimred_Valls_kernel}, etc., are also employed to handle non-linearity in HSIs. However, extracted features in the reduced dimensional space may not be optimal for classification. 
HSI classification task gets more complicated with the following facts: i) spectral signature of objects belonging to the same class may be different, and ii) the spectral signature of objects belonging to different classes may be the same. Therefore, only spectral components may not be sufficient to provide discriminating features for classification. Recent studies prove that incorporation of spatial context along with spectral components improves classification task considerably. 
There are two ways of exploiting spatial and spectral information for classification. In the first approach, spatial and spectral information are processed separately and then combined at the decision level~\cite{2S-Fusion,spatial-spectral_Lowrank}. The second strategy uses joint representation of spectral-spatial features~\cite{DR-CNN, DPP-ML, CNN-PPF, bassnet_Santra16}. In this paper, a novel joint representation of spectral-spatial features has been proposed. Our technique supersedes the accuracy of classification compared to the state of the art techniques.
%we have adopted the second strategy to accomplish the task of classification of hyperspectral images with higher accuracy than the state of the art techniques.
\par In the literature, 1-D~\cite{CNN-PPF}, 2-D~\cite{DR-CNN}, and 3-D~\cite{spatial-spectral_3dcnn_Zhang}
CNN based architectures are well-known for \emph{HSI} classification. Also,  hybridisation of different CNN-type architecture is  employed~\cite{spatial-spectral_3d_2d_HybridSN}. 1D-CNNs use pixel-pair strategy~\cite{CNN-PPF} which combines a set of pairs of training samples. Each pair is constructed using all permutations of training pixels under consideration. This set not only reveals the neighborhood of the observed pixel but also increases the training samples. Yet, it can not use full power of spatial information in hyper-spectral image classification. It completely ignores neighborhood profile. In general, 2-D and 3-D CNN based approaches are more suitable in such a scenario. However, there are many other architectures, e.g., \emph{Deep Belief Network}~\cite{DBN1, DBN2, DBN3, DBN4}, Autoencoders~\cite{autoencoder1,  autoencoder3, autoencoder4, autoencoder5} which provide efficient solution to the hyperspectral image classification problem. Multi-scale convolution is widely popular in literature for exploring complex spatial context. In most cases, outputs of different kernels are either concatenated~\cite{multi-scale-cnn,multi-scale-cnn2} or summed~\cite{ DR-CNN} together and further processed for feature extraction and classification. However, there exists no study on whether concatenating a large number of filter banks of different scales result in optimum classification.
In the present context, we are more interested in scrutinizing various CNN architectures for the current problem. In general, a few core components are available for making any CNN architecture. For example, convolution,  pooling, batch-normalization~\cite{batch_normalization}, and activation layers.  In practice, there are various ways of using convolution mechanism. A few of them are very popular, namely,  point convolution, group convolution, depth-wise separable convolution~\cite{separable convolution}, etc. Similarly, there is a variation to the pooling mechanism, namely, adaptive pooling~\cite{adaptive_pooling}. Recently, many mid-level components are developed, e.g., an inception module which integrates outputs of multi-scale convolutions. Mid-level components are sequentially combined to make a large network, such as, VGG~\cite{vgg}, GoogleNet~\cite{google}, etc. Additionally, a skip architecture~\cite{resnet} proves to be a successful way of making a very deep network to deal with the vanishing gradient problem. Hyperspectral image classification is still an interesting and challenging problem where the effectiveness of various core components of CNN  and their arrangement to resolve the classification problem, needs to be studied. To summarize, our work is motivated by the following observations.
\begin{enumerate}
	\item Incorporating spatial and spectral feature may achieve better classification accuracy for hyperspectral images. Hence, a strategy is needed to combine spatial and spectral information.
	
	\item Due to complex reflectance property in HSI, pixels of same class may have different spectral signature and pixels of different class may have similar spectral signature. Spatial neighborhood plays a key role in improving classification accuracies in such scenario.
	However, the neighborhood of a pixel at class boundary appears different compared to the non-boundary pixels. It is observed that non-boundary pixels include more pixels from the same class. However, the neighborhood of boundary pixels include pixels of different classes. Our strategy aims to bring similar neighborhood for the boundary and non-boundary pixels which belong to the same class.
	
	\item Substantial spectral information brings redundancy, and also taking all spectral bands together decreases the performance of classification. Hence, band reduction is required in hyperspectral image classification.
	
	\item A suitable network architecture is needed to handle the problems stated above such that the network can be trained in an end to end manner.

\end{enumerate}

In this paper, we present a \emph{CNN} architecture which performs three major tasks in a pipeline of processing, such as, :1) band reduction, 2) feature extraction, and 3) classification. The first block of processing uses point-wise 3-D convolution layers. For feature extraction, we use multi-scale convolution module.  In our work, we propose two architectures for feature extraction which eventually lead to two different \emph{CNN} architectures.
In the first architecture, we use an inception module with four parallel convolution structures for feature extraction.
Additionally, we use similar multi-scale convolutions in inception like structure but with a different arrangement. The second architecture extracts finer contextual information compared to the first one. We feed the extracted features to a fully connected layer to form the high-level representation. We train our network in an end to end manner by minimizing cross-entropy loss using Adam optimizer. Our proposed architecture gives a state of the art performance without any data augmentation on three benchmark HSI classification datasets. Besides this new architecture, we  propose a way to incorporate spatial information along with the spectral one. It not only covers the neighborhood profile for a given window, but also it observes the change of neighborhood by shifting its current window. We observe that this process appears to be more beneficial towards the boundary location compared to the a single window neighborhood system.
The contribution of this paper can be summarized as follows:
\begin{enumerate}
	\item A novel technique to obtain a joint representation of spatial and spectral features has been proposed. The design is aimed at improving classification accuracy at the boundary location of each class.
	
	\item A novel end to end shallow and wider neural network has been proposed, which is a hybridization of 3-D CNN with 2-D-CNN. This hybrid structure
	does band reduction followed by a discriminating feature extraction. Also, we have shown two different arrangements of similar multi-scale convolutional layers to extract distinctive features. 
\end{enumerate}
 
\Cref{sec:TPO} gives a detailed description of the proposed classification framework, including the technique of inclusion of the spatial information.  Performance and comparisons among the proposed networks and current state of the art methods are presented in \Cref{sec:experiments}. This paper is concluded in \Cref{sec:conclusion}.

%data augmentation
%introducing spatial usefullness
%may be there is only pixel pair but not patch -->need to discuss
%pixel pair means 1d channel but covering k*k patch

%network architechture

%same scales with different arrangements.
%\section{Related Work}
 %\subsection{Inception-Model}
 
% \subsection{Residual Network}
 %\subsection{Filter Groups}

%Moreover, samples from
%different classes usually present similar features. Therefore,
%the learned models for hyperspectral image representation tend
%to be redundant, where different factors focus on capturing
%similar features from the image. This leads to the decrease of
%the classification performance.

\section{PROPOSED CLASSIFICATION FRAMEWORK}
\begin{figure}
	\includegraphics[width=0.45\textwidth,trim=4 4 4 4,clip]{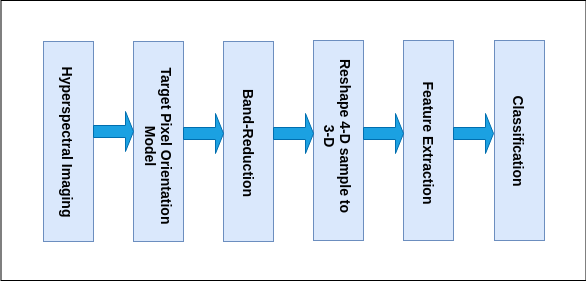}
	\caption{Flowchart of the proposed classification framework}
	\label{fig:Flow_diagram}
\end{figure}

The proposed classification framework shown in Fig~\ref{fig:Flow_diagram} mainly consists of three tasks: i) organizing a target-pixel-orientation model using available training samples, ii) constructing a CNN architecture to extract uncorrelated spectral information, and iii) learning spatial correlation with neighboring pixels.
%\IEEEpubidadjcol
\begin{figure}
	\centering
	\includegraphics[width=0.45\textwidth,trim=4 4 4 4,clip]{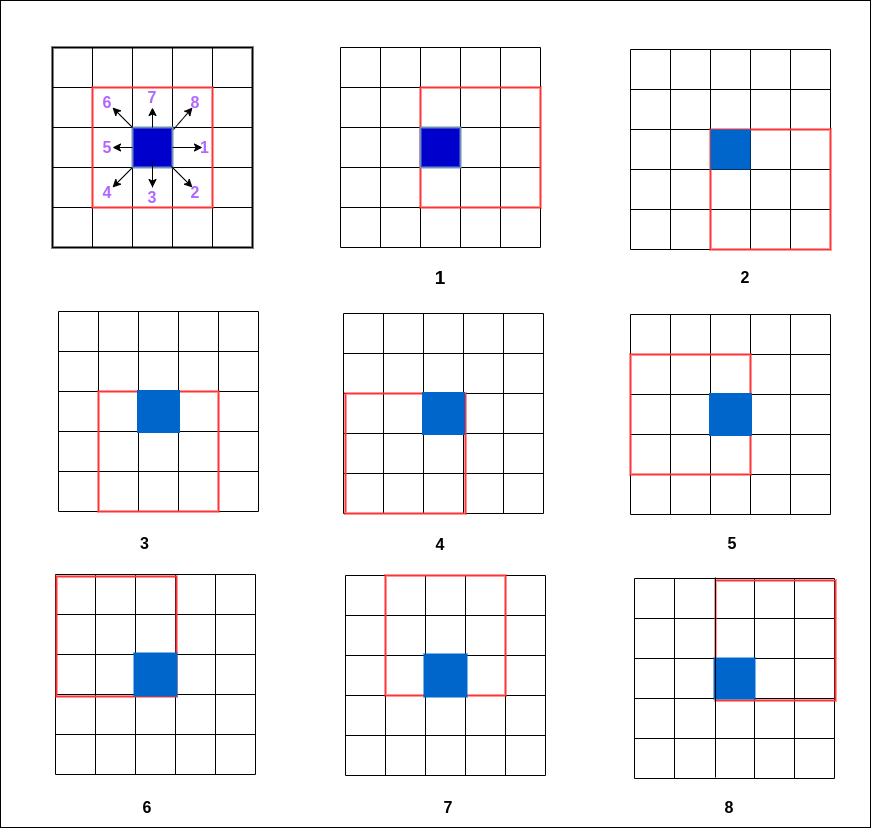}
	\caption{Example of Target-Patch-Orientation Model}
	\label{fig:TPO}
\end{figure}
\subsection{Target-Pixel-Orientation Model for Training Samples} \label{sec:TPO}
Consider a hyperspectral data set $H$ with $d$ spectral bands. We have $N$ labeled samples denoted as $P=\{p_i\}_{i=1}^{N}$ in an $R^{d \times 1}$ feature space and class labels $y_i \in \{1,2, \cdots, C\}$ where $C$ is the number of class.
Let $n_{c}$ be the number of available labeled samples in the $c^{th}$ class and $\sum_{c=1}^{C} n_c=N$. 
We propose a Target-Pixel-Orientation (TPO) scheme. In this scheme, we
consider a $k \times k$ window whose center pixel is the target pixel.
We select eight neighbors of the target pixel by simply shifting
the window into eight different directions in a clockwise manner. Fig~\ref{fig:TPO} shows one example of how we prepare eight neighboring of a target pixel of size 3x3. We mark the target pixel in blue color which is surrounded by a $3 \times 3$ neighboring window shown in a red border. First sub-image in Fig~\ref{fig:TPO} depicts the $3 \times 3$ window when the target pixel occupies the center position of that window. Other eight sub-images are the neighbors of the first sub-image which are numbered by 1 to 8. We consider each of nine windows as a single integrated view for the target pixel. However, we have shown the \emph{TPO} view with one spectral channel to make the illustration simple. In our proposed system, we  consider $d$ spectral channels. Therefore, input to the model is a 4-dimensional tensor. We perform the following operation to form the input for our models.
\begin{equation}
\mathbb{S}(\mathbb{S}(s_{ji}) \forall j\in d ) \forall i \in V
\end{equation}
Where $\mathbb{S}$ is a function which is responsible for stacking of channels and $s_{ji}$ represents $k \times k$ patch of $j^{th}$ spectral channel in $i^{th}$ view. $V$ represents nine views in the \emph{TPO} scheme.
We have converted labeled samples $P$ to $I=\{i_1,i_2,\cdots, i_N\}$ such that each $i_x$ has $9 \times d \times k \times k$ dimension.
%The reason is that the spatial
%information of the given pixel has positive effects on
%discriminating pixels from different classes. Moreover, it is
%worthwhile to note that the pixels near the border usually
%cannot be separated perfectly by spectral–spatial methods.
\subsubsection{Advantage of TPO for class boundary}
\begin{figure}
	\centering \includegraphics[width=0.45\linewidth]{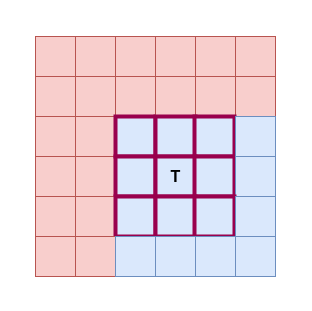} \qquad
	\includegraphics[width=0.45\linewidth,trim=4 4 4 4,clip]{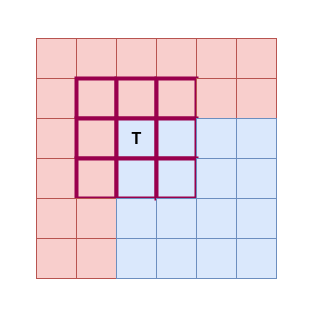}
	\caption{The position of a target pixel in a near-boundary and boundary position for a two-class scenario.}
	\label{fig:pixels}
%\end{figure}		
%\begin{figure}
	\centering
	\includegraphics[width=0.45\textwidth,trim=4 4 4 4,clip]{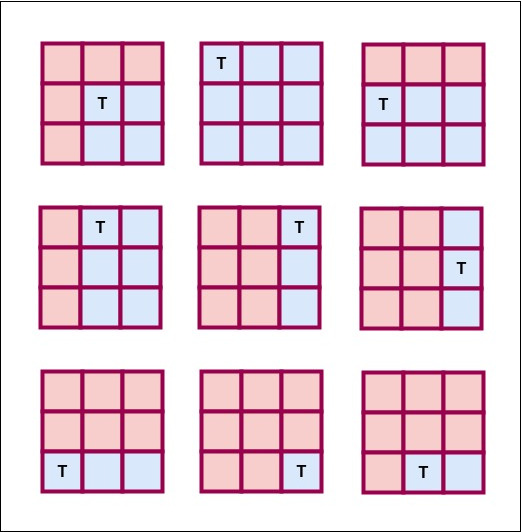}
	\caption{Example of Target-Patch-Orientation of a target pixel lies at the boundary of a class.}
	\label{fig:TPO_B}
%\end{figure}	
		
%\begin{figure}
	\centering
	\includegraphics[width=0.45\textwidth,trim=4 4 4 4,clip]{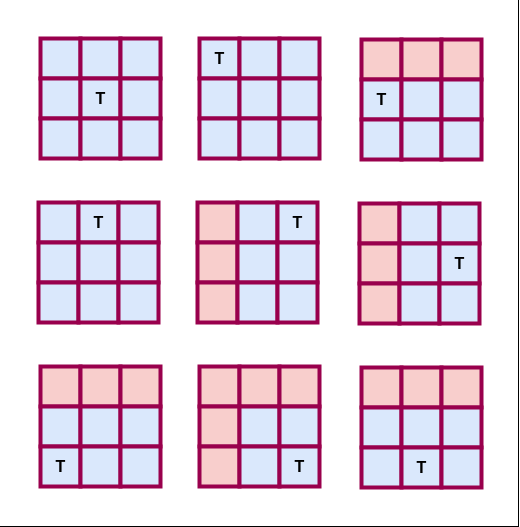}
	\caption{Example of Target-Patch-Orientation of a target pixel lies at the near boundary of a class.}
	\label{fig:TPO_NB}
\end{figure}

We observe that a $k \times k$ patch of a pixel appears very differently at the boundary region of any class compared to pixels in the non-boundary area.  In general, the non-boundary pixels are surrounded by pixels belonging to the same class. Also, all pixels in each view of the \emph{TPO} contains the same class information. However, the neighborhood of boundary pixels is contaminated with more than one class information. In such a scenario, \emph{TPO} provides nine 2d neighborhoods for the target pixel. Therefore, a few neighborhoods of boundary pixels are similar to the pixel at non-boundary regions.  The intuition is that the same neighborhoods may form a single cluster.  We illustrate this with a two-class situation in Fig-\ref{fig:pixels}. The patch of a target pixel at near boundary contains all pixels of similar class (blue). However, the patch of a target pixel at the border includes pixels of two classes (blue and red). If we consider only one patch surrounded that target pixel, we may fail to classify border pixels. In this scenario \emph{TPO} brings a different view of patches for a single target pixel at the boundary. We have shown \emph{TPO} of target pixel at border and near-border in Fig~\ref{fig:TPO_B} and Fig~\ref{fig:TPO_NB} respectively. In the given situation, there is at least one view where every pixel belongs to the blue class for the border pixel. However, there are other views which are similar to the views of the pixel at the near boundary. 

\subsection{Network Architecture}
The framework of the HSI classification is shown in Fig~\ref{fig:Flow_diagram}. It consists of mainly three blocks, namely, band-reduction, feature extraction, and classification. 
\emph{TPO} extracts samples from the given dataset as described in \Cref{sec:TPO}. The label of each sample is that of the pixel located in the center of the first view among the nine views (discussed in Sec~\ref{sec:TPO}). 
\subsubsection{Band-Reduction} \label{sec:pqr}

\begin{figure}
	\centering
	\includegraphics[width=0.45\textwidth,trim=4 4 4 4,clip]{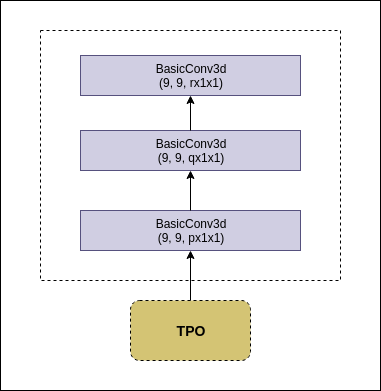}
	\caption{Diagram of the Band Reduction layer in the proposed network.}
	\label{fig:band_reduction-1}
\end{figure}
This block contains three consecutive ``BasicConv3d" layers. The designed ``BasicConv3d” layer contains 3-D batch-normalization layer and rectified linear unit (ReLU) layer sequentially after 3-D point-wise convolution layer. Parameters of 3-D convolution layer are the input channel, output channel, and the kernel size. In our experiment we have kept input channel$=9$ and output channel$=9$. However, we have empirically adjusted kernel size of three ``BasicConv3d” layers which is of the form (X,1,1). Hence, we have used X=p, X=q, and X=r notation in defining kernel size in Fig~\ref{fig:band_reduction-1}. The aim of choosing such kernel dimension is not to change the spatial size but to reduce the number of bands.
Dimension of the input to this layer is  $v \times d \times k \times k$  where $v$ represents number of views in the \emph{TPO} scheme, $d$ represents spectral dimensionality  and $k$ is spatial size. We consider $v=9,~ d=103,$ and $k=5$ to illustrate the network description. The first 3-D convolutional layer (C1) primarily filters the input with dimension $9 \times 103 \times 5 \times 5$  with kernel of size  $8 \times 1 \times 1$,  producing a $9 \times 96 \times 5 \times 5$ feature map. In this layer dimension of spectral channels gets reduced from 103 to 96.\\
As we have used point-wise 3-D convolution, there is no change in the spatial size of the sample. But, the size of the spectral channel is changed based on the value of p which is 8 in this example. The size of the spectral channel in the convolved sample can be computed using the following equation.
\begin{equation}
\floor*{\frac{W-K+2P}{S}}+1.
\end{equation}
Where $W$ represents the size of the spectral channel, which is $103$ in this case. $K, P$, and $S$ represent kernel size, padding, and stride.  For the above example, $K=8$, $P=0$ and $S=1$ holds. Therefore, we are getting 96 channels in the convolved sample.  The second layer (C2) combines the features obtained in the C1 layer with nine   $16 \times 1 \times 1$ kernels, resulting in a $9 \times 81 \times 5 \times 5$ feature map.  The third layer (C3) combines the features obtained in the C1 layer with nine $32 \times 1 \times 1$ kernels, resulting in a $9 \times 50 \times 5 \times 5$ feature map.  We have a reduced number of bands from 103 to 50 at this point. Now we reshape our data in 3 dimensions by stacking nine views for each 50 spectral information, leading to $450 \times 5 \times 5$-sized sample. We feed the reshaped output of band-reduction block to feature extraction layer.

\begin{figure}[htb!]
	\centering
	\includegraphics[width=0.5\textwidth, trim=4 4 4 4, clip]{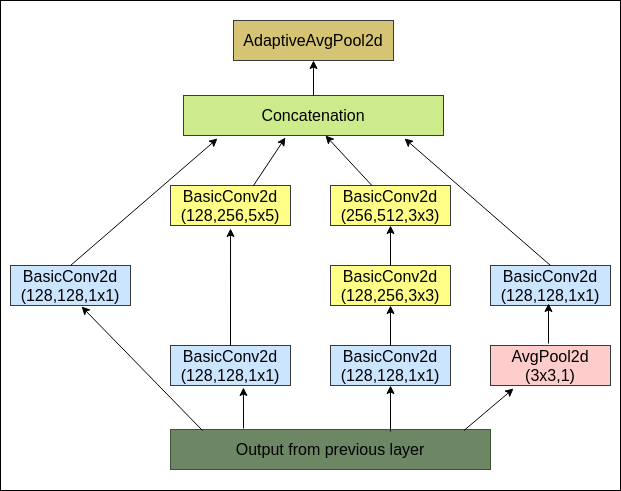}
	\caption{Diagram of the Feature Extraction layer (MS-CNN1) in the proposed network.}
	\label{fig:inception-1}
\end{figure}

\begin{figure*}[htb!]
	\centering
	\includegraphics[width=\textwidth,trim=4 4 4 4,clip]{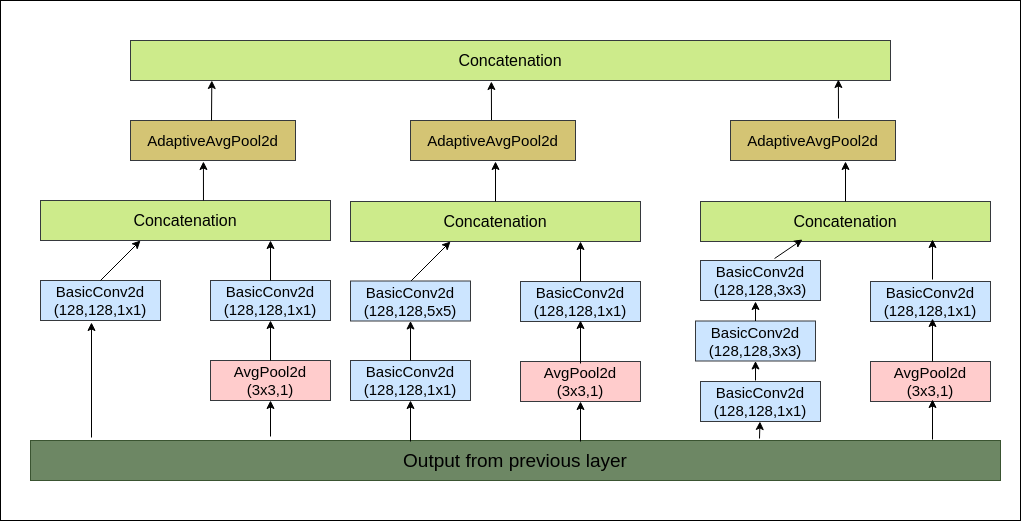}
	\caption{Diagram of the Feature Extraction layer (MS-CNN2) in the proposed network.}
	\label{fig:inception-2}
\end{figure*}

\subsubsection{Feature Extraction}
We have taken a tiny patch as an input sample. Our assumption is that a shallow but wider network, i.e., ``multi-scale filter bank" extracts more appropriate features from small patches. Hence, we have considered similar to inception-module for feature extraction. We use inception module in two different ways forming two separate networks. Fig~\ref{fig:inception-1} and Fig~\ref{fig:inception-2} depict feature extraction modules of \emph{MS-CNN1} and \emph{MS-CNN2} . Each ``BasicConv2d" layer in the figures contains a 2-D batch-normalization layer and rectified linear units (ReLU)  sequentially after 2-D convolution layer.  Parameter of the 2-D convolution layer is the input channel, output channel, and the kernel size. Each rectangular block of ``BasicConv2d" in the diagram contains parameters of the 2-D convolution layer. We denote this by $\mathbb{C}_{k_1\times k_1 \times B}$, where $k_1$ refers to the kernel size of the convolution layer and $B$ is the number of input channel. On the other hand, each block of ``AvgPool2d" depicts the kernel size and the stride value for the average pooling layer in the diagram. We denote this by $\mathbb{P}_{k_2\times k_2}$, where $k_2$ refers to the kernel size of the pooling layer.
\emph{MS-CNN1} uses a multi-scale filter bank that locally convolves the input sample with four parallel blocks with different filter sizes in convolution layer. Each parallel block consists of either one or many ``BasicConv2D" layer and pooling layer.  \emph{MS-CNN1} has the following details: 
: $\mathbb{C}_{1\times1\times B}$ , $\mathbb{C}_{1\times1\times B}$ followed by $\mathbb{C}_{3\times3\times B}$ followed by  $\mathbb{C}_{3\times3\times B}$,  $\mathbb{C}_{1\times1\times B}$ followed by  $\mathbb{C}_{5\times5\times B}$ and $\mathbb{P}_{3\times3}$ followed by $\mathbb{C}_{1\times1 \times B}$. %where B is the number of input channels and $\mathbb{C}$ and $\mathbb{P}$ represent ``BasicConv2d" and Pooling layers. 
The $3\times 3$ and $5\times 5$ filters are used to
exploit local spatial correlations of the input sample while the
$1 \times 1$  filters are used to address correlations among nine views and their respective spectral information. The outputs of the \emph{MS-CNN1} feature extraction layer are combined at concatenation layer to form a joint view-spatio-spectral feature map used as input to the subsequent layers. However, since the size of the feature maps from the four convolutional filters is different from each other,  we have padded the input feature with zeros to match the size of the output feature maps of each parallel blocks.  For example, we have padded input with 0, 1 and 2 zeros for  $1 \times 1$ , $3 \times 3$  and $5 \times 5$ filters, respectively.
In \emph{MS-CNN1}, we have used an adaptive average pooling~\cite{adaptive_pooling} layer sequentially after the concatenation layer. However, we have split the inception architecture of \emph{MS-CNN1} into three small inception layer. Each has two parallel convolutional layers. Each concatenation layer is followed by an adaptive average pooling layer. Finally, we concatenate all the pooled information.

\subsubsection{Classification}	
Outputs of feature extraction block  are flattened and fed to the fully connected layers whose output channel is the number of class. The fully connected layers is followed by 1-D Batch-Normalization layer and a softmax activation function.  In general, the classification layer can be defined as
\begin{equation}
p=softmax(BN(Wa+b))
\end{equation}

where $a$ is the input of the fully connected layer, and $W$ 
and $b$ are the weights and bias of the fully connected layer,
respectively. BN($\cdot$) is the 1-D Batch-Normalization layer. $p$ is the
$C$-dimensional vector which represents the probability that a sample belongs to the $c^{th}$ class.

\subsection{Learning the Proposed Network}
We have trained the proposed networks by minimizing cross-entropy loss function. Let $Y=\{y_i\}_{i=1}^b$ represents the ground-truth for the training samples present in a batch  $b$. $P=\{p_{ic}\}_{i=1}^{b}$ denotes the conditional probability distribution of the model. The model predicts that $i^{th}$ training sample belongs to the $c^{th}$ class with probability $p_{ic}$. The cross-entropy loss function $\mathbb{L}_{cross-entropy}$ is given by
\begin{equation}
\mathbb{L}_{cross-entropy}=-\sum_{i=1}^{b}\sum_{c=1}^{C}y_i[c]\log{p_{ic}}
\end{equation}
In our dataset, the ground truth is represented as one-hot encoded vector. Each $y_i$ is a $C$ dimensional vector and $C$ represents the number of classes.
If the class label of the $i^{th}$ sample is $c$ then,
\[ \begin{cases} 
y_{i}[j]=1 & \text{if}~j=c \\
y_{i}[j]=0 & \text{, Otherwise} 
\end{cases}
\]
To train the model, Adam optimizer with a batch size of 512 samples is used with a weight decay of 0.0001 . We initially set a base
learning rate as 0.0001. All the layers are initialized from a uniform
distribution.

\section{Experimental Results}\label{sec:experiments}
\begin{figure}
	\centering
	\includegraphics[width=0.15\textwidth, trim=4 4 4 4, clip]{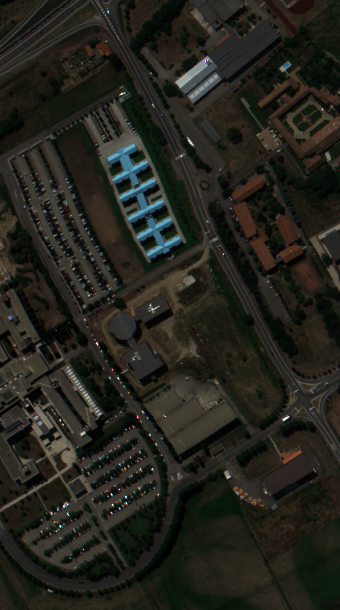} %\qquad
	\includegraphics[width=0.15\textwidth, trim=4 4 4 4, clip]{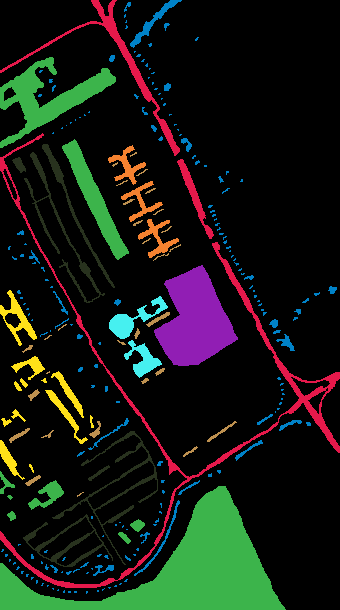}
	%\qquad
	\includegraphics[width=0.15\textwidth, trim=4 4 4 4, clip]{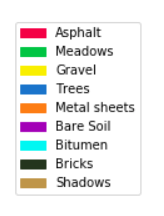}
	
	\caption{University of Pavia dataset. Three-band color composite image is given on left, ground truth is given in the middle, and color code used in ground-truth is shown on the right.}
	\label{fig:FC_gt_pavia}
%\end{figure}

%\begin{figure}
	\centering
	\includegraphics[width=0.15\textwidth, trim=4 4 4 4, clip]{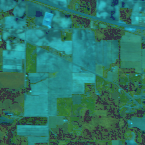} %\qquad
	\includegraphics[width=0.15\textwidth, trim=4 4 4 4, clip]{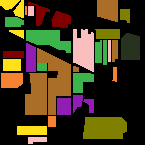}
	%\qquad
	\includegraphics[width=0.13\textwidth, trim=4 4 4 4, clip]{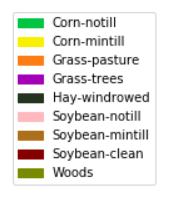}
	
	\caption{Indian Pines dataset. Three-band color composite image is given on left, ground truth is given in the middle, and color code used in ground-truth is shown on the right.}
	\label{fig:FC_gt_indian_pines}
%\end{figure}

%\begin{figure}
	\centering
	\includegraphics[width=0.15\textwidth, trim=4 4 4 4, clip]{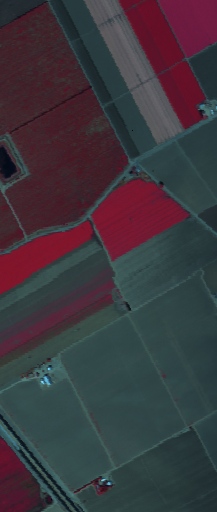} %\qquad
	\includegraphics[width=0.15\textwidth, trim=4 4 4 4, clip]{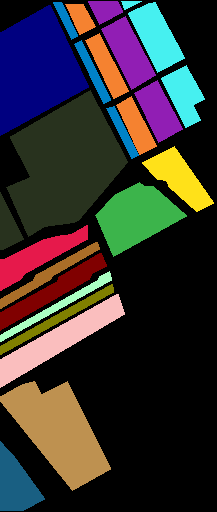}
	%\qquad
	\includegraphics[width=0.15\textwidth, trim=4 4 4 4, clip]{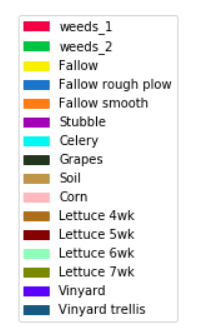}
	
	\caption{Salinas dataset. Three-band color composite image is given on left, ground truth is given in the middle, and color code used in ground-truth is shown on the right.}
	\label{fig:FC_gt_salinas}
\end{figure}

\begin{table}[htb!]
	\caption{A brief description on HSI Datasets}
	\begin{tabular}{llll}
		%\cline{2-4}
		\multicolumn{1}{l}{} & U. Pavia & Indian Pines & \multicolumn{1}{l}{Salinas} \\ \hline
		Sensor & ROSIS & AVIRIS & AVIRIS \\
		Place & \begin{tabular}[c]{@{}l@{}}Pavia, Northern\\ Italy\end{tabular} & \begin{tabular}[c]{@{}l@{}}Northwestern\\ Indiana\end{tabular} & \begin{tabular}[c]{@{}l@{}}Salinas valey\\ california\end{tabular} \\
		wavelength range & 0.43-0.86 $\mu m$ & 0.4-2.5 $\mu m$ & 0.4-2.5 $\mu m$ \\
		Spatial Resolution & 1.3 m & 20 m & 3.7 m \\
		No of bands & 103 & 220 & 224 \\
		No. of Classes & 9 & 16 & 16 \\ 
		Image size & $610 \times 340$ & $145 \times 145$ & $512 \times 217$\\\hline
	\end{tabular}
\label{tab:dataset}
\end{table}
\subsection{Datasets}
The performance of \emph{HSI} classification is observed by experimenting with three popular datasets: the Pavia University scene (U.P)~(Fig-\ref{fig:FC_gt_pavia}), the Indian Pines (I.P)~(Fig-\ref{fig:FC_gt_indian_pines}), and the Salinas (S) dataset~(\ref{fig:FC_gt_salinas}). \Cref{tab:dataset} contains a brief description about the datasets. We have discarded $20$ water absorption bands in Indian Pines. Also,  we have rejected some classes in \emph{Indian Pines} dataset which has less than 400 samples.  We have selected 200 labeled pixels from each class  to prepare a training set for each of the three HSI datasets.
The rest of the labeled samples constitute the test set. As different spectral channels range differently, we normalize them to the range [0, 1] using the  function $\mathbb{F}(\cdot)$ defined in Eq.~\ref{eq:norm}, where $x$ denotes the pixel value of a given spectral channel and $\mu$ and $\sigma$ provide mean and standard deviation of the complete dataset.
\begin{equation}
\mathbb{F}(x)=\frac{x-\mu}{\sigma}
\label{eq:norm}
\end{equation}

\subsection{Quantitative Metrics}
We evaluate the performance of the proposed architecture quantitatively in terms of the following metrics.
\subsubsection{Overall Accuracy(OA)}
Overall Accuracy is computed using the following formulae in the test samples where $C=$number of classes, considered for a given HSI dataset:
\begin{equation}
OA=\frac{\sum_{c_i \in C}\text{\#Correctly classified samples in class~} c_i}{\sum_{c_i \in C}\text{\#samples in class~} c_i}
\end{equation}
\subsubsection{Average Accuracy (AA)}
Average Accuracy is computed using the following formulae in the test samples where $C=$number of classes, considered for a given HSI dataset:
\begin{equation}
AA=\frac{1}{C}\sum_{c_i\in C}\frac{ \text{\#correctly classified samples in class}~c_i}{ \text{\#samples in class}~c_i }
\end{equation}

\subsubsection{$\kappa$-score}
The $\kappa$-score~\cite{kappa} is a statistical measure about the agreement between two classifiers. Each classifier classifies N samples into C mutually exclusive classes. $\kappa$-score is given by the following equation:

\begin{equation}
\kappa=\frac{p_o-p_e}{1-p_e}
\end{equation}

where $p_o$ is the relative observed agreement between classifiers, and $p_e$ is the hypothetical probability of chance agreement. $\kappa=1$ indicates complete agreement between two classifiers, while $\kappa\leq 0$ refers no agreement at all.

\subsection{Implementation Platform}
The network is implemented in pytorch\footnote{\url{https://pytorch.org/}}, a popular deep learning library, written in python. We have trained our models on a machine with \emph{GeForce GTX 1080 Ti GPU}.
\subsection{Comparison with Other Methods}
The key features of our proposed methods are 1) \emph{use of} both spectral and spatial features, 2) {\emph band-reduction} using several consecutive 3-D CNNs and 3) feature extraction with a {\emph multi-scale convolutional} network. We have chosen six state of the art methods namely,: 1) \emph{CNN-PPF}~\cite{CNN-PPF}, 2) \emph{DR-CNN}~\cite{DR-CNN}, 3)  \emph{2S-Fusion}~\cite{2S-Fusion}, 4) \emph{BASS}~\cite{bassnet_Santra16}, 5) \emph{DPP-ML}~\cite{DPP-ML}, and 6) \emph{S-CNN+SVM}~\cite{SVM-SCNN}. Every comparable method exploits both spectral and spatial features.  \emph{CNN-PPF } uses a pixel pair strategy to increase the number of training samples and feeds them into a deep network having 1-D convolutional layers.  \emph{DR-CNN} exploits diverse region-based 2-D patches from the image to produce more discriminative features. On the contrary,  \emph{2S-Fusion} processes spatial and spectral information separately and fuses them using adaptive class-specific weights. However, \emph{BASSNET} extracts band specific spatial-spectral features. In \emph{DPP-ML}, convolutional neural networks 
with multiscale convolution are used to extract deep multiscale features from the
HSI. SVM-based methods are common in traditional hyperspectral image classification.  In \emph{S-CNN+SVM}, the Siamese convolutional neural network extracts spectral-spatial features of HSI and feeds them to a \emph{SVM} classifier. In general, the performance of deep learning-based algorithms supersedes traditional techniques (e.g, \emph{k-NN}, \emph{SVM}, \emph{ELM}).
We have compared the performance of the proposed techniques with the best results reported for each of these state of the art techniques. In \emph{S-CNN+SVM} and \emph{2S-Fusion}, performance on the \emph{salinas} dataset is not reported. %We restrict ourselves from implementing their work due to absence of various parameters of the models in their paper.
To maintain consistency in the results, we ran our algorithm with the classes and the number of samples for each class used in \emph{2S-Fusion}, \emph{DR-CNN}, \emph{DPP-ML} for Indian Pines.

%incorporated similar classes of a given dataset and keep training samples from each class to 200.  Otherwise, we have implemented their work with the same architecture, learning algorithm, and hyperparameter values, as mentioned in the original paper. 

\begin{table*}[htb!]
	\centering
	\captionsetup{justification=centering}
	\caption{Class-specific Accuracy(\%) and OA of comparable techniques for the University of Pavia dataset }
	%\scalebox{0.8}{%
	\begin{tabular}{cccccccccc}
		\hline
		{Class}  & 	\multicolumn{1}{l}{\begin{tabular}[c]{@{}l@{}}Training \\samples\end{tabular}}& \multicolumn{1}{l}{\begin{tabular}[c]{@{}l@{}}CNN\\ -PPF\end{tabular}} & \multicolumn{1}{l}{BASS} & \multicolumn{1}{l}{\begin{tabular}[c]{@{}l@{}}S-CNN\\ +SVM\end{tabular}} & \multicolumn{1}{l}{\begin{tabular}[c]{@{}l@{}}2S\\ -Fusion\end{tabular}} & \multicolumn{1}{l}{\begin{tabular}[c]{@{}l@{}}DR\\ -CNN\end{tabular}} & \multicolumn{1}{l}{\begin{tabular}[c]{@{}l@{}}DPP\\ -ML\end{tabular}} & \multicolumn{1}{l}{\begin{tabular}[c]{@{}l@{}}MS-\\ CNN1\end{tabular}} & \multicolumn{1}{l}{\begin{tabular}[c]{@{}l@{}}MS-\\ CNN2\end{tabular}} \\ \hline
		Asphalt & 200 & 97.42 & 97.71 & \textbf{100} & 97.47 & 98.43 & 99.38 & 99.78 & \textbf{100} \\
		Meadows & 200 & 95.76 & 97.93 & 98.12 & 99.92 & 99.45 & 99.59 & 99.88 & \textbf{99.99} \\
		Gravel & 200 & 94.05 & 94.95 & 99.12 & 83.80 & 99.14 & 97.33 & 99.21 & \textbf{100} \\
		Trees & 200 & 97.52 & 97.80 & 99.40 & 98.98 & 99.50 & 99.31 & 99.41 & \textbf{99.93} \\
		Painted metal sheets & 200 & \textbf{100} & \textbf{100} & 99.18 & \textbf{100} & \textbf{100} & \textbf{100} & \textbf{100} & \textbf{100} \\
		Bare Soil & 200 & 99.13 & 96.60 & 99.10 & 97.75 & \textbf{100} & 99.99 & 99.75 & \textbf{100} \\
		Bitumen& 200  & 96.19 & 98.14 & 98.50 & 77.44 & 99.70 & 99.85 & \textbf{100} & \textbf{100} \\
		Self-Blocking Bricks& 200  & 93.62 & 95.46 & 99.91 & 96.65 & 99.55 & 99.02 & 99.77 & \textbf{100} \\
		Shadows& 200  & 99.60 & \textbf{100} & \textbf{100} & 99.65 & \textbf{100} & \textbf{100} & \textbf{100} & \textbf{100} \\ \hline
		OA  &  & 96.48 & 99.68 & 97.50 & 99.56 & 99.46 & 99.72 & 99.78&\textbf{99.99} \\ \hline
	\end{tabular}
\label{tab:pavia_OA}
\end{table*}

\begin{table}[htb!]
	\centering
	%\captionsetup{justification=left}
	\caption{Class-specific Accuracy(\%) and OA of comparable techniques for the Indian Pines dataset. Training and testing is restricted to 9 classes.}
%\scalebox{0.8}{	
\begin{tabular}{cccccc}
	\hline
	\multicolumn{6}{c}{Number of training samples per class=200}\\\hline
	Class  & \begin{tabular}[c]{@{}c@{}}CNN-\\ PPF\end{tabular} & BASS & \begin{tabular}[c]{@{}c@{}}S-CNN\\ +SVM\end{tabular} & \begin{tabular}[c]{@{}c@{}}MS-\\ CNN1\end{tabular} & \begin{tabular}[c]{@{}c@{}}MS-\\ CNN2\end{tabular} \\ \hline
	
	Corn-notil  & 92.99 & 96.09 & 98.25 & \textbf{100} & \textbf{100} \\
	Corn-mintil  & 96.66 & 98.25 & 99.64 & \textbf{99.92} & 99.75 \\
	Grass-pasture  & 95.58 & \textbf{100} & 97.10 & \textbf{100} & 99.68\\
	Grass-tree  & \textbf{100} & 99.24 & 99.86 & 99.73 & 99.82 \\
	Hay-windrowed  & \textbf{100} & \textbf{100} & \textbf{100} & \textbf{100} & \textbf{100} \\
	Soybean-notil  & 96.24 & 94.82 & 98.87 & \textbf{100} & \textbf{100} \\
	Soybean-mintil  & 87.80 & 94.41 & 98.57 & 99.74 & \textbf{100} \\
	Soybean-clean & 98.98 & 97.46 & \textbf{100} & \textbf{100} & \textbf{100} \\
	Woods  & 99.81 & 99.90 & \textbf{100} & \textbf{100} & 99.72 \\ \hline
	\multicolumn{1}{c}{OA} & 94.34 & 96.77 & 99.04 & \textbf{99.89} & 99.84 \\ \hline
\end{tabular}
%}
\label{tab:indian_pines_OA}
\end{table}

\begin{table*}[htb!]
	\centering
	\captionsetup{justification=centering}
	\caption{Class-specific Accuracy(\%) and OA of comparable techniques for the Indian Pines dataset. DR-CNN and DPP-ML consider 8 classes and 2S-Fusion includes 16 classes for the experiment. }
	%\scalebox{0.8}{
\begin{tabular}{c|ccccc|cccc}
	\hline
	\multicolumn{1}{c|}{Class}&\multicolumn{1}{l}{\begin{tabular}[c]{@{}l@{}} Training \\samples\end{tabular}} & \multicolumn{1}{l}{\begin{tabular}[c]{@{}l@{}} DR\\-CNN\end{tabular}} & \multicolumn{1}{l}{\begin{tabular}[c]{@{}l@{}}DPP\\-ML\end{tabular}} & \multicolumn{1}{l}{\begin{tabular}[c]{@{}l@{}}MS-\\ CNN1\end{tabular}} & \multicolumn{1}{l|}{\begin{tabular}[c]{@{}l@{}}MS-\\ CNN2\end{tabular}} &
	\multicolumn{1}{l}{\begin{tabular}[c]{@{}l@{}} Training \\samples\end{tabular}}& \multicolumn{1}{l}{\begin{tabular}[c]{@{}l@{}}2S-\\ Fusion\end{tabular}} & \multicolumn{1}{l}{\begin{tabular}[c]{@{}l@{}}MS-\\ CNN1\end{tabular}} & \multicolumn{1}{l}{\begin{tabular}[c]{@{}l@{}}MS-\\ CNN2\end{tabular}} \\ \hline
	Alfalfa& -& - & - & - & - &33 &\textbf{100} & \textbf{100} & \textbf{100} \\
	Corn-notill&200 & 98.20 & 99.03 & \textbf{100} & \textbf{100} &861 &95.35 & \textbf{100} & \textbf{100} \\
	Corn-mintill&200 & \textbf{99.79} & 99.74 & 99.51 & 99.67 & 501&98.75 & \textbf{100} & \textbf{100} \\
	Corn &- & - & - & - & - &141 &\textbf{100} & \textbf{100} & \textbf{100} \\
	Grass-pasture&200 & \textbf{100} & \textbf{100} & \textbf{100} & \textbf{100} & 299&\textbf{100} & \textbf{100} & \textbf{100} \\
	Grass-trees&-& - & - & - & - & 449&99.32 & \textbf{100} & \textbf{100} \\
	Grass-pasture-mowed&- & - & - & - & - & 9&\textbf{100} & \textbf{100} & \textbf{100} \\
	Hay-windrowed &200& \textbf{100} & \textbf{100} & 98.84 & 98.85 & 294&\textbf{100} & \textbf{100} & \textbf{100} \\
	Oats &-& - & - & - & - & 12&\textbf{100} & \textbf{100} & \textbf{100} \\
	Soybean-notill&200 & 99.78 & 99.61 & \textbf{100} & \textbf{100} & 580&\textbf{100} & \textbf{100} & \textbf{100} \\
	Soybean-mintill&200 & 96.69 & 97.80 & \textbf{100} & 99.91 & 1480&98.03 & \textbf{100} & \textbf{100} \\
	Soybean-clean&200 & 99.86 & \textbf{100} & \textbf{100} & \textbf{100} & 369&\textbf{100} & \textbf{100} & \textbf{100} \\
	Wheat&- & - & - & - & - & 127&97.87 & \textbf{100} & \textbf{100} \\
	Woods&200 & 99.99 & \textbf{100} & \textbf{100} & \textbf{100} & 777&99.62 & \textbf{100} & \textbf{100} \\
	Buildings-Grass-Trees-Drives&- & - & - & - & - & 228&98.53 & \textbf{100} & \textbf{100} \\
	Stone-Steel-Towers&- & - & - & - & - &57 &\textbf{100} & \textbf{100} & \textbf{100} \\ \hline
	OA && 98.54 & 99.08 & 99.54 & \textbf{99.55} & &98.65 & \textbf{100} & \textbf{100} \\
	%AA & 99.29 & 99.52 & 99.79 & 99.80 & 99.22 & 100 & 100 \\ 
	\hline
\end{tabular}
%}
\label{tab:indian_pines_other}
\end{table*}

\begin{table*}[htb!]
	\centering
	\captionsetup{justification=centering}
	\caption{Class-specific Accuracy(\%) and OA of comparable techniques for the Salinas dataset }
	%\scalebox{0.9}{%
	\begin{tabular}{ccccccccc}
	\hline
	\multicolumn{1}{c}{Class}&\multicolumn{1}{l}{\begin{tabular}[c]{@{}l@{}} Training \\samples\end{tabular}}  & \multicolumn{1}{l}{\begin{tabular}[c]{@{}l@{}}CNN\\ -PPF\end{tabular}} & \multicolumn{1}{l}{BASS} & %\multicolumn{1}{l}{\begin{tabular}[c]{@{}l@{}}S-CNN\\ +SVM\end{tabular}} &
	% \multicolumn{1}{l}{\begin{tabular}[c]{@{}l@{}}2S\\ -Fusion\end{tabular}} &
	\multicolumn{1}{l}{\begin{tabular}[c]{@{}l@{}}DR\\ -CNN\end{tabular}} & \multicolumn{1}{l}{\begin{tabular}[c]{@{}l@{}}DPP\\ -ML\end{tabular}} & \multicolumn{1}{l}{\begin{tabular}[c]{@{}l@{}}MS-\\ CNN1\end{tabular}} & \multicolumn{1}{l}{\begin{tabular}[c]{@{}l@{}}MS-\\ CNN2\end{tabular}} \\ \hline
	Brocoli-green-weeds-1&200  & \textbf{100} & \textbf{100} & \textbf{100} & \textbf{100} & \textbf{100} & \textbf{100} \\
	Brocoli-green-weeds-2&200  & 99.88 & 99.97  & 100 & \textbf{100} & \textbf{100} & \textbf{100} \\
	Fallow&200 & 99.60 & \textbf{100}  & 99.98 & \textbf{100} & \textbf{100} & 99.72 \\
	Fallow-rough-plow&200  & 99.49 & 99.66  & 99.89 & 99.25 & \textbf{100} & \textbf{100} \\
	Fallow-smooth&200  & 98.34 & 99.59  & 99.83 & 99.44 & 99.84 & \textbf{99.88} \\
	Stubble&200  & 99.97 & 100  & \textbf{100} & \textbf{100} & \textbf{100} & \textbf{100} \\
	Celery&200 & \textbf{100} & 99.91  & 99.96 & 99.87 & \textbf{100} & \textbf{100} \\
	Grapes-untrained&200  & 88.68 & 90.11 & 94.14 & 95.36 & 94.30 & \textbf{98.17} \\
	Soil-vinyard-develop&200  & 98.33 & 99.73 & 99.99 & \textbf{100} & 99.75& \textbf{100} \\
	Corn-senesced-green-weeds&200  & 98.60 & 97.46 & 99.20 & 98.85 & 94.02 & 99.35 \\
	Lettuce-romaine-4wk&200 & 99.54 & 99.08  & 99.99 & 99.77 & \textbf{100} & \textbf{100} \\
	Lettuce-romaine-5wk&200  & \textbf{100} & \textbf{100} & \textbf{100} & \textbf{100} & \textbf{100} & \textbf{100} \\
	Lettuce-romaine-6wk&200 & 99.44 & 99.44  & \textbf{100} & 99.86 & \textbf{100} & \textbf{100} \\
	Lettuce-romaine-7wk&200  & 98.96 & \textbf{100} & \textbf{100} & 99.77 & \textbf{100} & \textbf{100} \\
	Vinyard-untrained&200  & 83.53 & 83.94 & \textbf{95.52} & 90.50 & 95.08 & 94.03 \\
	Vinyard-vertical-trellis&200  & 99.31 & 99.38 & 99.72 & 98.94 & \textbf{100} & \textbf{100} \\ \hline
	OA & & 94.80 & 95.36 & 98.33 & 97.51 & 97.98 & \textbf{98.72} \\ \hline
\end{tabular}
%}
\label{tab:salinas_OA}
\end{table*}

\subsection{Results and Discussion}
The performance of the proposed \emph{MS-CNN1} and \emph{MS-CNN2} on test-samples are compared with the aforementioned deep learning-based classifiers in \Cref{tab:pavia_OA,tab:indian_pines_OA,tab:indian_pines_other,tab:salinas_OA}.  
We have considered  the size of spatial window ($k$) for generating a patch as $7$ to evaluate the performance of our algorithms.
For \emph{Indian-Pines}, \emph{DR-CNN} and \emph{DPP-ML} used 8 classes,   \emph{2S-Fusion} used 16 classes and other comparable methods used 9 classes for their experiment. To compare our method we have chosen the same number of classes and training samples of each class as mentioned in the respective papers. \Cref{tab:indian_pines_OA} and \Cref{tab:indian_pines_other}  together indicate that MS-CNN1 and MS-CNN2 provides comparable results and supersedes the other methods. 
 However, \emph{MS-CNN2} produces better results compared to \emph{MS-CNN1} in \emph{University of Pavia} and \emph{Salinas} datasets as shown in \Cref{tab:pavia_OA} and \Cref{tab:salinas_OA} respectively. The results signify that the arrangements of multi-scale convolutions in \emph{MS-CNN2} is able to extract more useful features for the classification compared to \emph{MS-CNN1}.

\begin{figure*}
	\centering
	\begin{tabular}[l]{llll}
		\centering
		\begin{subfigure}[b]{0.23\textwidth}
			\includegraphics[width=1.2\textwidth]{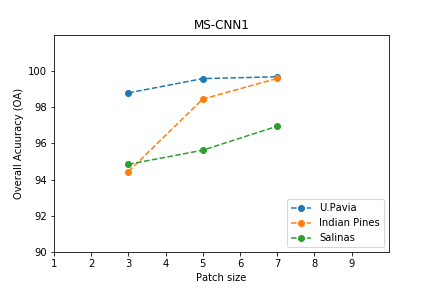}
			\caption{}
			\label{fig:param_ss1}
		\end{subfigure}&
		\begin{subfigure}[b]{0.23\textwidth}
			\includegraphics[width=1.2\textwidth]{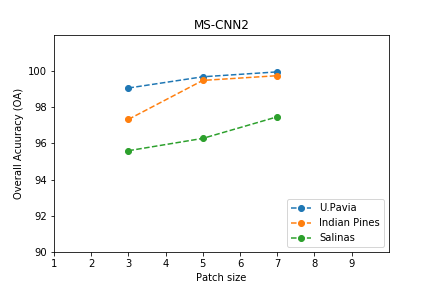}
			\caption{}
			\label{fig:param_ss3}
		\end{subfigure}&
		\begin{subfigure}[b]{0.23\textwidth}
			\includegraphics[width=1.2\textwidth]{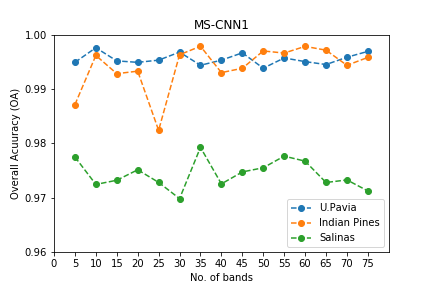}
			\caption{}
			\label{fig:param_r1}
		\end{subfigure}&
		\begin{subfigure}[b]{0.23\textwidth}
			\includegraphics[width=1.2\textwidth]{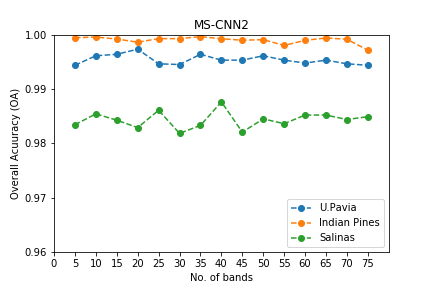}
			\caption{}
			\label{fig:param_r2}
		\end{subfigure}
		
	\end{tabular}
	\caption{Variation of test accuracy on the three HSI datasets (a)-(b) with varying patch size, (c)-(d) with reduced number of spectral channels.}
	
\end{figure*}

\begin{figure*}
	\centering
	\begin{tabular}[l]{llllll}
		\centering
		\begin{subfigure}[b]{0.14\textwidth}
			\includegraphics[width=\textwidth]{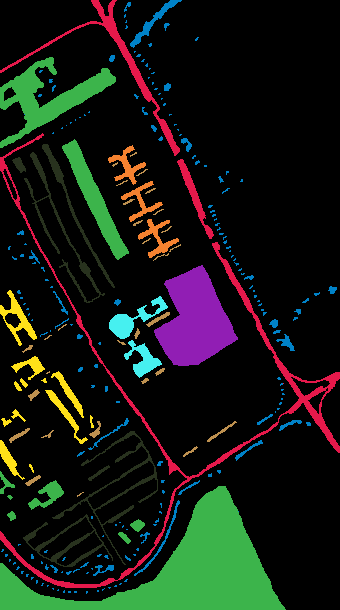}
			\caption{}
			%\label{fig:waterbody_s1}
		\end{subfigure}&
		\begin{subfigure}[b]{0.14\textwidth}
			\includegraphics[width=\textwidth]{images/paviaU_TPO_CNN2_7x7.png}
			\caption{}
			%\label{fig:built-up_s1}
		\end{subfigure}&
		
		\begin{subfigure}[b]{0.107\textwidth}
			\includegraphics[width=\textwidth]{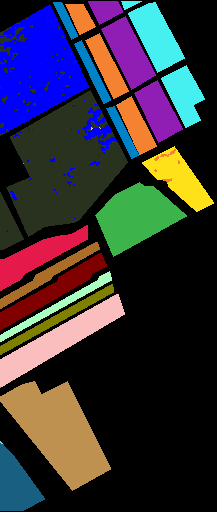}
			\caption{}
			%\label{fig:waterbody_s1}
		\end{subfigure}&
		\begin{subfigure}[b]{0.107\textwidth}
			\includegraphics[width=\textwidth]{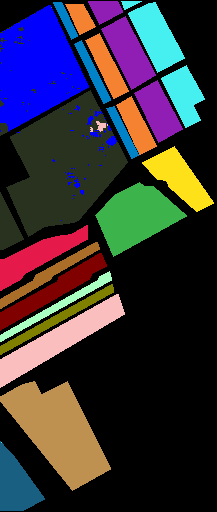}
			\caption{}
			%\label{fig:built-up_s1}
		\end{subfigure}&

\begin{subfigure}[b]{0.15\textwidth}
	\includegraphics[width=\textwidth]{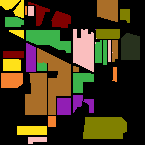}
	\caption{}
	%\label{fig:waterbody_s1}
\end{subfigure}&
\begin{subfigure}[b]{0.15\textwidth}
	\includegraphics[width=\textwidth]{images/indian_pines_TPO_CNN2_7x7.png}
	\caption{}
	%\label{fig:built-up_s1}
\end{subfigure}
	
	\end{tabular}
	\caption{Thematic maps resulting from classification using $7 \times 7$-patch by  MS-CNN1 and MS-CNN2 respectively for (a)-(b) University of Pavia dataset, (c)-(d) Salinas data set, and (e)-(f) Indian Pines dataset. Color code is similar to its ground truths.}
	\label{fig:thematic_map}
\end{figure*}

We have shown thematic maps generating from the classification of three \emph{HSI} scenes using our networks in \Cref{fig:thematic_map}.  In order to check the consistency of our network, we repeat experiments 10 times with different training sets. \Cref{tab:mean} shows the mean and standard deviation of OA over these 10 experiments for each data set. %\emph{TPO-CNN2} outperforms all the other methods on all the three datasets in terms of OA of classification. However, \emph{TPO-CNN1} outperforms other compared methods only in \emph{indian pines} dataset. %We have adopted the paired t-test at 95\% significance level to illustrate that the \emph{TPO-CNN2} is statistically better than the baselines (i.e, \emph{S-CNN+SVM} and \emph{DR-CNN}).

\begin{table}[htb!]
		\centering
	\caption{Mean and standard deviation of 10 Independent Experiments}
	\label{tab:mean}

\begin{tabular}{clcccccc}
	\hline
	\multicolumn{2}{c}{Datasets} & \multicolumn{2}{c}{U.P} & \multicolumn{2}{c}{I.P} & \multicolumn{2}{c}{S} \\ \hline
	\multicolumn{2}{c}{MS-CNN} & 1 & 2 & 1 & 2 & 1 & 2 \\ \hline
	\multirow{2}{*}{OA} & Mean & 99.76 & 99.90 & 99.67 & 99.65 & 98.10 & 98.65 \\
	& Std-dev & 0.24 & 0.06 & 0.14 & 0.17 & 0.37 & 0.28 \\ \hline
	\multirow{2}{*}{AA} & Mean & 99.80 & 99.94 & 99.82 & 99.78 & 97.87 & 98.49 \\
	& Std-dev & 0.20 & 0.06 & 0.07 & 0.11 & 0.42 & 0.30 \\ \hline
	\multicolumn{1}{l}{\multirow{2}{*}{$\kappa$}} & Mean & 99.44 & 99.86 & 99.61 & 99.58 & 99.29 & 99.44 \\
	\multicolumn{1}{l}{} & Std-dev & 0.32 & 0.08 & 0.17 & 0.20 & 0.18 & 0.14 \\ \hline
\end{tabular}

\end{table}

\begin{table}[htb!]
	\centering
	\caption{Overall Accuracy (OA) for varying patch size}
		%\scalebox{0.8}{%
\begin{tabular}{|c|ccc|ccc|}
	\hline
	\multicolumn{1}{|c|}{\multirow{2}{*}{Patch size}} & \multicolumn{3}{c|}{MS-CNN1} & \multicolumn{3}{c|}{MS-CNN2} \\ \cline{2-7} 
	\multicolumn{1}{|c|}{} & \multicolumn{1}{c}{U.P} & \multicolumn{1}{c}{I.P} & \multicolumn{1}{c|}{S} & \multicolumn{1}{c}{U.P} & \multicolumn{1}{c}{I.P} & \multicolumn{1}{c|}{S} \\ \hline
	$3\times3$ & 98.78 & 94.44 & 94.84 & 99.05 & 97.32 & 95.59 \\
	$5\times5$ & 99.57 & 98.44 & 95.62 & 99.67 & 99.47 & 96.27 \\
	$7\times7$ & 99.67 & 99.58 & 96.94 & 99.94 & 99.73 &  97.46\\\hline
	%$9\times9$ & 99.92 & 99.78 &  & 99.96 & 99.85 &  \\ \hline
\end{tabular}
%}

%	\begin{tabular}{l|lll|ccc}
%		\hline
%		\multirow{2}{*}{Datasets} & \multicolumn{3}{c|}{TPO-CNN1} & \multicolumn{3}{c}{TPO-CNN2} \\ \cline{2-7} 
%		& $3 \times 3$ & $5 \times 5$ & $7 \times 7$ & $3 \times 3$ & $5 \times 5$ & $7 \times 7$ \\ \hline
%		U.Pavia & 98.7 & 99.7 & 99.7 & 98.9 & 99.9 & 99.9 \\
%		Indian Pines & 96.9 & 98.5 & 99.7 & 97.2 & 99.5 & 99.8 \\
%		Salinas & 91.1 & 96.0 & 93.7 & 93.8 & 96.7 & 98.8 \\ \hline
%	\end{tabular}
\end{table}

\subsection{Comparison of Different Hyper-parameter Settings}
There are two hyper-parameters which have a direct effect on the accuracy of classification task: 1) Spatial size of input image, and the  2) Number of spectral channels obtained from the band reduction block. \Cref{fig:param_ss1,fig:param_ss3} depicts test accuracies on 3 \emph{HSI} datasets  for different choices of input patch size on the same randomly selected training sample. We observe that with increasing patch size accuracies in \emph{University of Pavia}, \emph{Indian Pines} and \emph{Salinas} increase in both the networks. %This behavior again supports the fact that the arrangements of multi-scale convolutions of \emph{MS-CNN2} is superior to \emph{MS-CNN1} with respect to feature extraction.
\Cref{tab:param_br} shows the adjusted $(p, q, r)$ (refer to \Cref{sec:pqr}) parameters used for 3 \emph{HSI} datasets. We vary the value of $r$ to get a different number of bands and observe its impact on classification accuracy. We did not observe any monotonically increasing or decreasing behavior in overall classification accuracy for changing the number of bands with varying value of $r$. \Cref{fig:param_r1,fig:param_r2} depicts the change in overall  accuracy (OA) for varying $r$. They suggest that ($r_{\text{MS-CNN2}}=35$ and $r_{\text{MS-CNN1}}=35$), ($r_{\text{MS-CNN2}}=20$ and $r_{\text{MS-CNN1}}=10$) ,($r_{\text{MS-CNN2}}=40$ and $r_{\text{MS-CNN1}}=35$) are suitable for \emph{University of Pavia}, \emph{Indian Pines}, \emph{Salinas}, respectively.  

\begin{table}[htb!]
	\centering
	\caption{Parameters in Band Reduction Block}
	\begin{tabular}{l|ccc|}
		\cline{2-4}
		& \multicolumn{1}{l}{U.P} & \multicolumn{1}{l}{I.P} & \multicolumn{1}{l|}{S} \\ \hline
		\multicolumn{1}{|l|}{p} & 8 & 32 & 32 \\
		\multicolumn{1}{|l|}{q} & 16 & 57 & 61 \\
		\multicolumn{1}{|l|}{r} & 32 & 64 & 64 \\ \hline
	\end{tabular}
\label{tab:param_br}
\end{table}

\subsection{Classification performance with decreasing number of training samples}

In this section, the influence of decreasing the number of samples on the classification accuracy has been studied on the \emph{University of Pavia}, \emph{Indian Pines}, and \emph{Salinas} datasets.  We present the experimental results with the setup as mentioned above with $7 \times 7$ spatial resolution. Here, each class selects a fixed number $(t)$ of samples per class from the labeled pixels. To showcase the effect of decreasing number of training samples on the
classification accuracy, we have chosen several values of $t$, e.g., 150, 100, and 50.
Our proposed architecture still performs better than  most of the comparable methods with 150 training samples per class. \Cref{tab:small_sample} reassures that network architecture for feature extraction in \emph{MS-CNN2} brings more useful feature for \emph{University of Pavia}, \emph{India Pines}, and \emph{Salinas} datasets even with a small number of training samples compared to \emph{MS-CNN1}. %However we observe a small deviation of this results in the \emph{Salinas} dataset with 100 training samples per class.

\begin{table}[htb!]
	\centering
	\caption{Performance measures on the three datasets with decreasing number of training samples. metrics are the average of 10 independent experiments.}
	\begin{tabular}{cccccccc}
		\hline
		\multicolumn{2}{c}{\#samples} & \multicolumn{2}{c}{150} & \multicolumn{2}{c}{100} & \multicolumn{2}{c}{50} \\ \hline
		\multicolumn{2}{c}{MS-CNN} & 1 & 2 & 1 & 2 & 1 & 2 \\ \hline
		\multirow{3}{*}{U.P} & \begin{tabular}[c]{@{}c@{}} OA \\ std. dev. \end{tabular} &\begin{tabular}[c]{@{}c@{}} 99.58 \\(0.22) \end{tabular} & \begin{tabular}[c]{@{}c@{}}99.93 \\(0.05) \end{tabular}& \begin{tabular}[c]{@{}c@{}}99.16\\ (0.62)\end{tabular} & \begin{tabular}[c]{@{}c@{}}99.75\\ (0.13)\end{tabular} & \begin{tabular}[c]{@{}c@{}}98.25\\ (1.15)\end{tabular} & \begin{tabular}[c]{@{}c@{}}98.98\\ (0.62)\end{tabular} \\
		& \begin{tabular}[c]{@{}c@{}} AA \\ std. dev. \end{tabular} & \begin{tabular}[c]{@{}c@{}} 99.76\\ (0.08)\end{tabular} & \begin{tabular}[c]{@{}c@{}} 99.92\\(0.06) \end{tabular} & \begin{tabular}[c]{@{}c@{}}98.88\\ (0.82)\end{tabular} & \begin{tabular}[c]{@{}c@{}}99.67\\ (0.17)\end{tabular} & \begin{tabular}[c]{@{}c@{}}97.68\\ (1.50)\end{tabular} & \begin{tabular}[c]{@{}c@{}}98.65\\ (0.82)\end{tabular} \\
		& \begin{tabular}[c]{@{}c@{}} $\kappa$ \\ std. dev. \end{tabular} & \begin{tabular}[c]{@{}c@{}} 99.44\\(0.30) \end{tabular} & \begin{tabular}[c]{@{}c@{}} 99.91\\ (0.07)\end{tabular} & \begin{tabular}[c]{@{}c@{}}99.45\\ (0.31)\end{tabular} & \begin{tabular}[c]{@{}c@{}}99.82\\ (0.07)\end{tabular} & \begin{tabular}[c]{@{}c@{}}98.54\\ (0.74)\end{tabular} & \begin{tabular}[c]{@{}c@{}}99.17\\ (0.39)\end{tabular} \\ \hline
		\multirow{3}{*}{\begin{tabular}[c]{@{}c@{}}I.P\end{tabular}} & \begin{tabular}[c]{@{}c@{}} OA \\ std. dev. \end{tabular} & \begin{tabular}[c]{@{}c@{}}99.44\\ (0.12)\end{tabular} & \begin{tabular}[c]{@{}c@{}}99.47\\ (0.18)\end{tabular} & \begin{tabular}[c]{@{}c@{}}98.72\\ (0.46)\end{tabular} & \begin{tabular}[c]{@{}c@{}}98.88\\ (0.35)\end{tabular} & \begin{tabular}[c]{@{}c@{}}94.12\\ (1.32)\end{tabular} & \begin{tabular}[c]{@{}c@{}}95.88\\ (1.02)\end{tabular} \\
		& \begin{tabular}[c]{@{}c@{}} AA \\ std. dev. \end{tabular} & \begin{tabular}[c]{@{}c@{}}99.33\\ (0.15)\end{tabular} & \begin{tabular}[c]{@{}c@{}}99.38\\ (0.22)\end{tabular} & \begin{tabular}[c]{@{}c@{}}98.49\\ (0.55)\end{tabular} & \begin{tabular}[c]{@{}c@{}}98.67\\ (0.42)\end{tabular} & \begin{tabular}[c]{@{}c@{}}93.10\\ (1.54)\end{tabular} & \begin{tabular}[c]{@{}c@{}}95.22\\ (1.25)\end{tabular} \\
		& \begin{tabular}[c]{@{}c@{}} $\kappa$ \\ std. dev. \end{tabular}  & \begin{tabular}[c]{@{}c@{}}99.59\\ (0.20)\end{tabular} & \begin{tabular}[c]{@{}c@{}}99.60\\ (0.24)\end{tabular} & \begin{tabular}[c]{@{}c@{}}99.15\\ (0.30)\end{tabular} & \begin{tabular}[c]{@{}c@{}}99..23\\ (0.19)\end{tabular} & \begin{tabular}[c]{@{}c@{}}96.21\\ (0.77)\end{tabular} & \begin{tabular}[c]{@{}c@{}}97.18\\ (0.50)\end{tabular} \\ \hline
		\multirow{3}{*}{S} & \begin{tabular}[c]{@{}c@{}} OA \\ std. dev. \end{tabular} & \begin{tabular}[c]{@{}c@{}}97.39\\ (0.24)\end{tabular} & \begin{tabular}[c]{@{}c@{}}97.51\\ (0.39)\end{tabular} & \begin{tabular}[c]{@{}c@{}}96.48\\ (0.44)\end{tabular} & \begin{tabular}[c]{@{}c@{}}96.92\\ (0.39)\end{tabular} & \begin{tabular}[c]{@{}c@{}}95.81\\(0.29) \end{tabular} & \begin{tabular}[c]{@{}c@{}}96.28\\ (0.39)\end{tabular} \\
		& \begin{tabular}[c]{@{}c@{}} AA \\ std. dev. \end{tabular} & \begin{tabular}[c]{@{}c@{}}99.06\\(0.11)\end{tabular} & \begin{tabular}[c]{@{}c@{}}99.21\\(0.09)\end{tabular} & \begin{tabular}[c]{@{}c@{}}98.60\\ (0.12)\end{tabular} & \begin{tabular}[c]{@{}c@{}}98.68\\ (0.20)\end{tabular} & \begin{tabular}[c]{@{}c@{}}98.26\\(0.33) \end{tabular} & \begin{tabular}[c]{@{}c@{}}98.48\\(0.24) \end{tabular} \\
		& \begin{tabular}[c]{@{}c@{}} $\kappa$ \\ std. dev. \end{tabular}  &\begin{tabular}[c]{@{}c@{}}97.08\\ (0.27)\end{tabular}  & \begin{tabular}[c]{@{}c@{}}97.22\\ (0.44)\end{tabular} & \begin{tabular}[c]{@{}c@{}}96.08\\ (0.49)\end{tabular} & \begin{tabular}[c]{@{}c@{}}96.56\\ (0.44)\end{tabular} & \begin{tabular}[c]{@{}c@{}}95.33\\(0.33) \end{tabular} & \begin{tabular}[c]{@{}c@{}}95.86\\(0.43) \end{tabular} \\ \hline
	\end{tabular}
\label{tab:small_sample}
\end{table}

\subsection{Ablation study of the TPO strategy}
\begin{figure*}[htb!]
	\centering
	\makebox[\linewidth]{
		\begin{tabular}[l]{llllll}
			\centering
			\begin{subfigure}[b]{0.14\textwidth}
				\includegraphics[width=\textwidth,height=0.25\textheight]{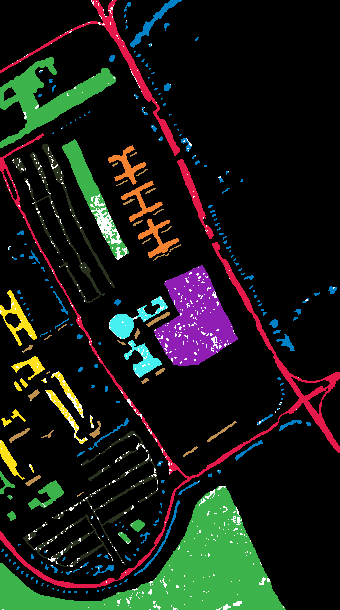}
				\centering
				\caption{without TPO}
				\label{fig:TPO1_paviau1}
			\end{subfigure}&
			\begin{subfigure}[b]{0.14\textwidth}
				\includegraphics[width=\textwidth,height=0.25\textheight]{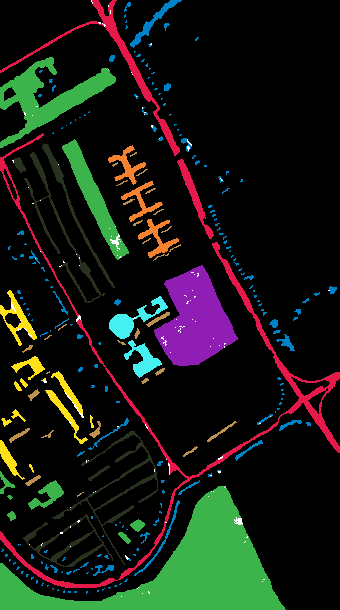}
				\centering
				\caption{with TPO}
				\label{fig:TPO1_paviau9}
			\end{subfigure}&
			\begin{subfigure}[b]{0.15\textwidth}
				\includegraphics[=\textwidth, height=0.25\textheight]{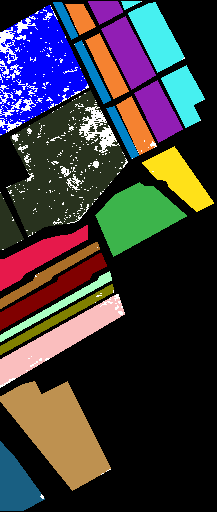}
				\centering
				\caption{without TPO}
				\label{fig:TPO1_salinas1}
			\end{subfigure}&
			\begin{subfigure}[b]{0.15\textwidth}
				\includegraphics[width=\textwidth, height=0.25\textheight]{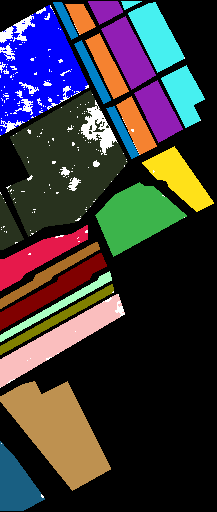}
				\centering
				\caption{with TPO}
				\label{fig:TPO1_salinas9}
			\end{subfigure}&
			
			\begin{subfigure}[b]{0.14\textwidth}
				\includegraphics[width=\textwidth]{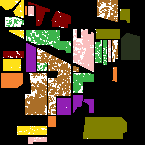}
				\centering
				\caption{without TPO}
				\label{fig:TPO1_ip1}
			\end{subfigure}&
			\begin{subfigure}[b]{0.14\textwidth}
				\includegraphics[width=\textwidth]{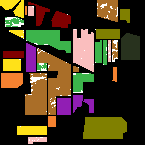}
				\centering
				\caption{with TPO}
				\label{fig:TPO1_ip9}
			\end{subfigure}		
			
		\end{tabular}	
	}
	\caption{Thematic maps resulting from classification using $3 \times 3$-patch by \emph{MS-CNN1} for  (a)-(b) University of Pavia dataset, (c)-(d) Salinas dataset  and (c)-(d) Indian Pines dataset respectively. Color code is similar to its ground truths. White color code is used to show misclassified pixels.}
	\label{fig:miscalssification_TPO1}
	
%\end{figure*}
%\begin{figure*}
%	\vspace{-0.85cm}
	\centering
	\makebox[\linewidth]{
		\begin{tabular}[l]{llllll}
			\centering
			\begin{subfigure}[b]{0.14\textwidth}
				\includegraphics[width=\textwidth, height=0.25\textheight]{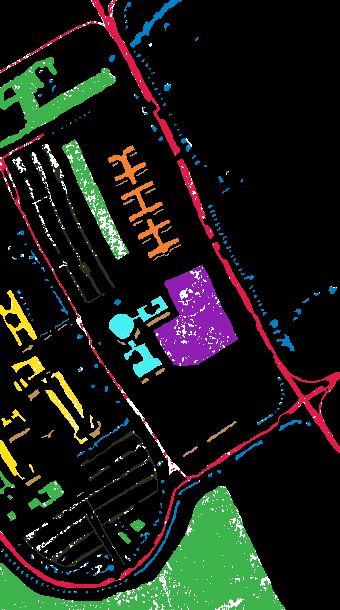}
				\centering
				\caption{without TPO}
				\label{fig:TPO2_paviau1}
			\end{subfigure}&
			\begin{subfigure}[b]{0.14\textwidth}
				\includegraphics[width=\textwidth,height=0.25\textheight]{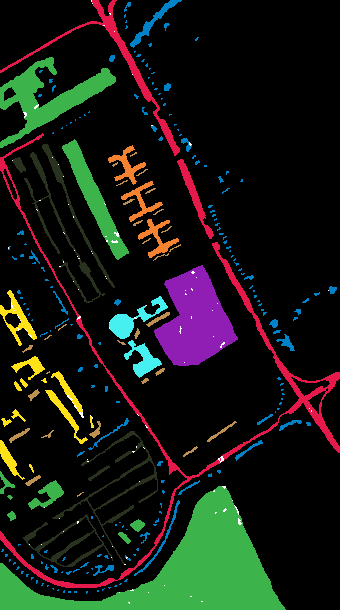}
				\centering
				\caption{with TPO}
				\label{fig:TPO2_paviau9}
			\end{subfigure}&
			\begin{subfigure}[b]{0.14\textwidth}
				\includegraphics[width=\textwidth,height=0.25\textheight]{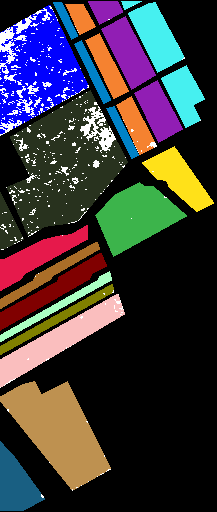}
				\centering
				\caption{without TPO}
				\label{fig:TPO2_salinas1}
			\end{subfigure}&
			\begin{subfigure}[b]{0.14\textwidth}
				\includegraphics[width=\textwidth, height=0.25\textheight]{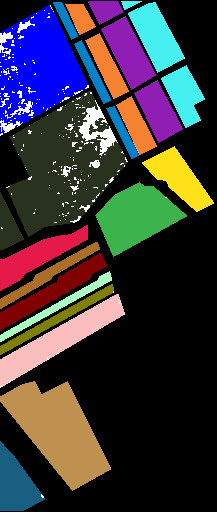}
				\centering
				\caption{with TPO}
				\label{fig:TPO2_salinas9}
			\end{subfigure}&
			
			\begin{subfigure}[b]{0.14\textwidth}
				\includegraphics[width=\textwidth]{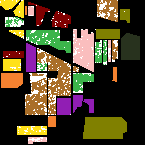}
				\centering
				\caption{without TPO}
				\label{fig:TPO2_ip1}
			\end{subfigure}&
			\begin{subfigure}[b]{0.14\textwidth}
				\includegraphics[width=\textwidth]{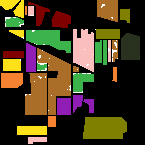}
				\centering
				\caption{with TPO}
				\label{fig:TPO2_ip9}
			\end{subfigure}		
		\end{tabular}
	}
	\caption{Thematic maps resulting from classification using $3 \times 3$-patch by \emph{MS-CNN2} for  (a)-(b) Salinas dataset , (c)-(d) University of Pavia dataset, and (e)-(f)  Indian Pines dataset respectively. Color code is similar to its ground truths. White color code is used to show misclassified pixels.}
	\label{fig:miscalssification_TPO2}	
\end{figure*}

In order to judge how the \emph{TPO} strategy, as described in \Cref{sec:TPO} affects the performance of the classifier, we compare the classification results using a single view where the position of the target pixel is the center of the given window. \Cref{tab:tpo} supports the fact that the \emph{TPO} strategy has a direct effect on the classification accuracies. We observe that OA increases by 6.09\%, 13.18\% 1.37\% in \emph{MS-CNN1} model and 9.99\%, 14.55\% 2.08\% in \emph{MS-CNN2} model for classification of \emph{University of Pavia}, \emph{Indian Pines}, \emph{Salinas}, respectively. In brief, the \emph{TPO}-scheme improves results  in every dataset for both the models. This suggests consistent behavior and a positive impact of \emph{TPO}-strategy on each model for all three datasets. Again, we observe \emph{MS-CNN2} provides better results compared to \emph{MS-CNN1}. We have highlighted misclassified pixels with white color code in \Cref{fig:miscalssification_TPO1} and \Cref{fig:miscalssification_TPO2}. 
\subsubsection{Description of metrics for quantitative analysis}
We have proposed a few metrics to analyze the impact of the TPO scheme in the models. $\tau$ in \Cref{eq:tau} reveals the percentage of misclassified pixels along the boundary (B) or non-boundary (NB) regions. 
\begin{equation}
\tau=\frac{\sum_{c_i\in C} \text{\#samples misclassified in B (NB)}}{\sum_{c_i\in C} \text{\#samples in B (NB)}} \times 100
\label{eq:tau}
\end{equation}
We have proposed $\nu(a \rightarrow b)$ in \Cref{eq:nu} to understand the impact of including TPO scheme in the proposed models. $tc(\cdot)$ determines truely classified pixels. $mc(\cdot)$ determines misclassified pixels. $a,b\in\{Yes, No\}$. The condition (TPO=a) indicates whether TPO scheme is considered.
\begin{equation}
\resizebox{\hsize}{!}{
	$\nu(a \rightarrow b)=\frac{\sum_{c_i\in C} \text{\#samples in B (NB)}~[mc(TPO=a)~ \text{and}~ tc(TPO=b)]}{\sum_{c_i\in C} \text{\#samples in B (NB)}} \times 100$
}
\label{eq:nu}
\end{equation}
 Similarly, We have proposed $\mu(a \rightarrow b)$ in \Cref{eq:mu} to understand the impact of switching models.  $\mathcal{M}$ denotes deep model whose values can be $a$ or $b$. $a,b\in\{\text{MS-CNN1}, \text{MS-CNN2}\}$
\begin{equation}
\label{eq:mu}
\resizebox{\hsize}{!}{
$\mu(a \rightarrow b)=\frac{\sum_{c_i\in C} \text{\#samples in B (NB)}  [~ mc(\mathcal{M}=a)~ \text{and}~  tc(\mathcal{M}=b)~]}{\sum_{c_i\in C} \text{\#samples in B (NB)}} \times 100
$}
\end{equation}
\subsubsection{Impact of TPO scheme on each  model}
\Cref{tab:tau_BNB} clearly shows that the misclassification rate reduces along the boundary and non-boundary regions with the inclusion of the TPO scheme.
\begin{figure}[htb!]
	\includegraphics[width=0.5\textwidth]{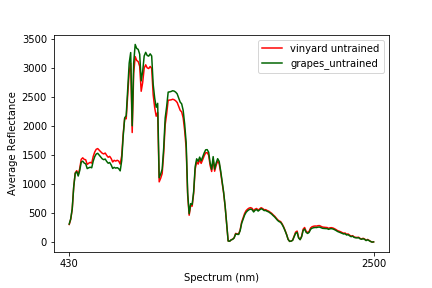}
	\caption{Average reflectance spectrum of  pixels whose prediction is similar to ground truth by each model with and without the \emph{TPO} scheme,}
	\label{fig:ARS}
\end{figure}

\begin{figure*}
	\centering
%	\makebox[\linewidth]{
%		\resizebox{\textwidth}{!}{
	\begin{tabular}[l]{ll}
		\centering
		\begin{subfigure}[b]{0.5\textwidth}
			\includegraphics[width=\textwidth]{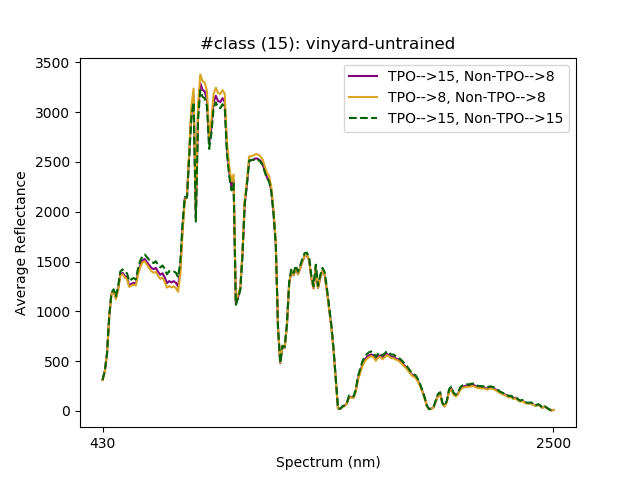}
			\caption{}
			\label{fig:built-up_s1}
		\end{subfigure}&	
		\begin{subfigure}[b]{0.5\textwidth}
			\includegraphics[width=\textwidth]{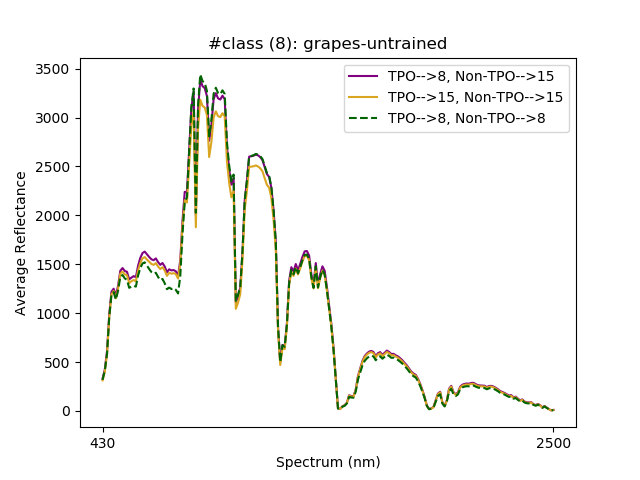}
			\caption{}
			\label{fig:waterbody_s1}
		\end{subfigure}
		
	\end{tabular}
%}}
	\caption{Average reflectance spectrum of selected samples based on three conditions. TPO$\rightarrow$`X' and Non-TPO$\rightarrow$`Y' represents prediction for the sample by the two models with the \emph{TPO} scheme is `X' and non-\emph{TPO} scheme is `Y'. Legends show the three conditions.}
	\label{fig:ARS_TPO}
\end{figure*}
A critical challenge in hyper-spectral imaging is that some pixels which belong to the same land cover class may have different spectral signatures due to complex light scattering mechanism. Therefore, an approach that is capable of making generalized features for the case, as mentioned above, can offer better classification accuracies. We discuss the impact of the TPO approach on the above issue in Salinas dataset. We found that the class=`vinyard-untrained' has highest percentage of misclassification by every model irrespective of TPO scheme. Also, it is identified that most of them are  misclassified as 'grapes-untrained'. %Hence, we have chosen this class for illustration. 
For `vinyard-untrained' and `grapes-untrained' in \emph{Salinas}, we calculate the average of reflectance of selected samples for each spectral band belonging to the respective class and
name it Average Reflectance Spectrum (ARS).
In \Cref{fig:ARS}, we have shown ARS for `vinyard-untrained' and `grapes-untrained' in \emph{Salinas} dataset . As shown in this figure, two of them  have very similar reflectance spectrum. In \Cref{fig:ARS_TPO}, we have selected samples for computing ARS on the basis of three conditions: i) samples that are correctly predicted by each model with \emph{TPO} and non-{TPO} scheme, ii) samples are correctly predicted by each model with the \emph{TPO} scheme but misclassified by each model with non-\emph{TPO} scheme, and iii) samples that are misclassified by each model with \emph{TPO} and non-{TPO} scheme. The first condition incorporates most of the samples of the respective class. \Cref{fig:ARS_TPO} clearly reveals that the class has different reflectance spectrum within itself. 
Additionally, it shows that the  \emph{TPO} scheme is able to predict correctly a fraction of class members whose reflectance spectrum is different from the majority of the class members. 
\Cref{eq:vinyard_grapes} quantitatively analyze the improvement in classification of pixels due to incorporation of the \emph{TPO} approach. We have tabulated values of \Cref{eq:vinyard_grapes} in \Cref{tab:vynyard_grapes}. It clearly states that \emph{TPO}-scheme has positive effect on the class which has different spectral signature. 
\begin{equation}
\label{eq:vinyard_grapes}
\resizebox{\hsize}{!}{
$\alpha(X|Y)=\frac{\text{\#samples in class 'Y' misclassifed as class 'X' by the two models in B (NB)}}{\text{\#samples of class 'Y' in B (NB)}}$
}
\end{equation}

Additionally, \Cref{tab:nu} reveals that the percentage of boundary pixels that are correctly classified with TPO inclusion is more compared to the non-boundary region in all three datasets. However, we have observed that a tiny fraction which is correctly classified without including TPO-scheme
gets misclassified with TPO-scheme.  

\subsubsection{Impact of models on TPO scheme }
\Cref{tab:mu} considers two cases. Case-1 denotes the number of pixels that are misclassified by \emph{MS-CNN1} but correctly classified by \emph{MS-CNN2}. Case-2 indicates the number of pixels that are misclassified by \emph{MS-CNN2} but correctly classified by \emph{MS-CNN1}. When case-1 supersedes case-2, we may conclude \emph{MS-CNN2} is a better performer compared to \emph{MS-CNN1}. \Cref{tab:mu} suggests an inconsistent behavior without the TPO scheme. It shows case-1 supersedes case-2 for \emph{Indian Pines} in the non-boundary region and salinas in the boundary region. However, with \emph{TPO}-scheme, case-1 always supersedes case-2 for all three datasets in both boundary and non-boundary regions. Our observation suggests that \emph{MS-CNN2} performs better with the \emph{TPO}-scheme. However, \emph{MS-CNN1} has no generalized behavior for all three datasets in either including the \emph{TPO}-scheme or excluding it.
%\subsection{Impact of stratified sampling on the TPO scheme}
\section{Conclusion}\label{sec:conclusion}
In this paper, a hybrid of 3-D and 2-D CNN-based network architecture is proposed for \emph{HSI} classification. We also propose a strategy (namely, target-pixel orientation (\emph{TPO})) to incorporate spatial and spectral information of \emph{HSI}. In general, classification accuracy degrades due to a class having different  spectral signatures in HSI. With incorporation of spectral and spatial information together, classification accuracy can be improved. But,  mis-classification at the boundary region still exists as the spatial neighborhood is different in a boundary region compared to a non-boundary region. Our approach attempts to take care of this limitation by using the orientation of the Target-pixel-view. Our architectural design of neural network exploits point-wise 3-D convolutions for band reduction whereas we adopt multi-scale 2-D inception like architecture for feature extraction. We have tested the granular arrangement of multi-scale convolutions in inception like architecture in \emph{MS-CNN2}.
We find  \emph{MS-CNN2} provides better results compared to \emph{MS-CNN1}. The experimental results with real hyperspectral images demonstrates the positive impact of including \emph{TPO} strategy. Also, the proposed work improves the performance of the classification accuracy compared to the state of the art methods even with a smaller number of training samples (For example, 150 samples per class). All the experimental results suggest that the arrangement of multi-scale convolutions in \emph{MS-CNN2} provides more useful features compared to \emph{MS-CNN1}. Experiments show that TPO-scheme is working better in \emph{MS-CNN2}.

\begin{table}[htb!]
	\centering
	\caption{Impact of TPO-strategy on the classification considering $3\times3$ spatial size}
	\label{tab:tpo}
	\begin{tabular}{ccccc}
		\hline
		TPO & metric & U.P & \begin{tabular}[c]{@{}c@{}}I.P\end{tabular} & S\\ \hline
			\multicolumn{5}{c}{\emph{MS-CNN1}} \\ \hline
		\multirow{3}{*}{No} & \emph{OA} & 92.69 & 81.26 & 93.47 \\
		& \emph{AA}& 92.32 & 88.40 & 97.14 \\
		& $\kappa$ & 90.24 & 77.82 & 92.69 \\\hline
		\multirow{3}{*}{Yes} & \emph{OA} & \textbf{98.78} & \textbf{94.44} & \textbf{94.84} \\
		&\emph{AA}& \textbf{98.60} & \textbf{96.92} & \textbf{97.73} \\
		& $\kappa$ & \textbf{98.36} & \textbf{93.36} & \textbf{94.22} \\ \hline
	\multicolumn{5}{c}{\emph{MS-CNN2}} \\ \hline
		\multirow{3}{*}{No} & \emph{OA}& 89.06 & 82.27 & 93.51 \\
		&\emph{AA} & 91.14 & 88.73 & 97.26 \\
		& $\kappa$ & 85.50 & 78.97 & 92.73 \\\hline
		\multirow{3}{*}{Yes} & \emph{OA} & \textbf{99.05} & \textbf{97.32} & \textbf{95.59} \\
		& \emph{AA} & \textbf{98.94} & \textbf{98.40} & \textbf{98.35} \\
		& $\kappa$ & \textbf{98.72} & \textbf{96.79} & \textbf{95.07}\\\hline
	\end{tabular}
\end{table}

\begin{table}[htb!]
	\centering
	\caption{Percentage of pixels misclassified ($\alpha(X|Y)$) by both MS-CNN1 and MS-CNN2 }
	\begin{tabular}{llcc}
		\hline
		\multirow{2}{*}{\begin{tabular}[c]{@{}l@{}}\#True \\ Class\end{tabular}} & TPO & boundary & non-bounday \\ \cline{2-4} 
		& \multicolumn{3}{l}{misclassified as `grapes-untrained' (X)} \\ \hline
		\multirow{2}{*}{\begin{tabular}[c]{@{}l@{}}`vinyard-\\untrained (Y)' \end{tabular}} & Yes & \textbf{14.48} & \textbf{8.06} \\
		& No & 19.54 & 13.27 \\ \hline
		& \multicolumn{3}{l}{misclassified as `vinyard-untrained' (X)} \\ \hline
		\multirow{2}{*}{\begin{tabular}[c]{@{}l@{}}grapes-\\untrained (Y) \end{tabular}} & Yes & \textbf{9.20} & \textbf{7.02} \\
		& No & 9.61 & 7.55\\\hline
	\end{tabular}
	\label{tab:vynyard_grapes}
\end{table}

\begin{table}[htb!]
	\centering
	\caption{Total misclassification (in \%) of pixels ($\tau$) in  boundary and non-boundary region. Bold fond is used to highlight lower misclassification. }
	\begin{tabular}{clllll}
		\hline
		\multirow{2}{*}{Dataset} & Models & \multicolumn{2}{c}{MS-CNN1} & \multicolumn{2}{c}{MS-CNN2} \\ \cline{2-6} 
		& TPO & \multicolumn{1}{c}{No} & \multicolumn{1}{c}{Yes} & \multicolumn{1}{c}{No} & \multicolumn{1}{c}{Yes} \\	\hline
		\multirow{2}{*}{U. P} & Boundary & \multicolumn{1}{c}{{8.03}} & \multicolumn{1}{c}{\textbf{1.75}} & \multicolumn{1}{c}{{12.39}} & \multicolumn{1}{c}{\textbf{1.42}} \\
		& Non-boundary & \multicolumn{1}{c}{{6.42}} & \multicolumn{1}{c}{\textbf{0.84}} & \multicolumn{1}{c}{{9.42}} & \multicolumn{1}{c}{\textbf{0.62}} \\\hline
		\multirow{2}{*}{I.P} & Boundary & {18.60} & \textbf{8.10} & {19.27} & \textbf{8.14} \\
		& Non-boundary & {13.87} & \textbf{3.22} & {12.54} & \textbf{1.12} \\\hline
		\multirow{2}{*}{S} & Boundary & {6.69} & \textbf{5.72} & {6.18} & \textbf{4.26} \\
		& Non-boundary & {6.09} & \textbf{4.80} & {6.12} & \textbf{4.15}\\\hline
	\end{tabular}
\label{tab:tau_BNB}
\end{table}
\begin{table}[]
	\centering
	\caption{Impact of TPO scheme (metric $\nu(a\rightarrow b)$) for the improvement of classification in  boundary and non-boundary regions. N abbreviates No and Y abbreviates Yes.}
	\makebox[\linewidth]{
		\resizebox{0.5\textwidth}{!}{
	\begin{tabular}{clcccc}
		\hline
		\multirow{2}{*}{Dataset} & Models & \multicolumn{2}{c}{\emph{MS-CNN1}} & \multicolumn{2}{c}{\emph{MS-CNN2}} \\ \cline{2-6} 
		&  & \multicolumn{1}{c}{$\nu(N\rightarrow Y)$} & \multicolumn{1}{c}{$\nu(Y\rightarrow N)$} & \multicolumn{1}{c}{$\nu(N\rightarrow Y)$} & \multicolumn{1}{c}{$\nu(Y \rightarrow N)$} \\\hline
		\multirow{2}{*}{U. P} & Boundary & \multicolumn{1}{c}{\textbf{7.28}} & \multicolumn{1}{c}{0.97} & \multicolumn{1}{c}{\textbf{11.70}} & \multicolumn{1}{c}{0.74} \\
		& Non-boundary & \multicolumn{1}{c}{\textbf{5.98}} & \multicolumn{1}{c}{0.41} & \multicolumn{1}{c}{\textbf{9.19}} & \multicolumn{1}{c}{0.39} \\\hline
		\multirow{2}{*}{I.P} & Boundary & \textbf{14.13} & 3.63 & \textbf{15.73} & 1.60 \\
		& Non-boundary & \textbf{11.35} & 0.70 & \textbf{11.82} & 0.39 \\\hline
		\multirow{2}{*}{S} & Boundary & \textbf{3.23} & 2.26 & \textbf{3.67} & 1.75 \\
		& Non-boundary & \textbf{2.98} & 1.70 & \textbf{3.61} & 1.64\\\hline
	\end{tabular}
}
}
\label{tab:nu}
\end{table}
\begin{table}[htb!]
	\centering
	\caption{Impact of models (metric $\mu$) on the improvement of classification in  boundary and non-boundary region.\\ $a=\text{\emph{MS-CNN1}}$ and $b=\text{\emph{MS-CNN2}}$ }
	\makebox[\linewidth]{
		\resizebox{0.5\textwidth}{!}{
	\begin{tabular}{clcccc}
		\hline
		\multirow{2}{*}{Dataset} & \emph{TPO} & \multicolumn{2}{c}{No} & \multicolumn{2}{c}{Yes} \\ \cline{2-6} 
		& Models & \multicolumn{1}{c}{$\mu(a \rightarrow b)$} & \multicolumn{1}{c}{$\mu(b \rightarrow a)$} & \multicolumn{1}{c}{$\mu(a \rightarrow b)$} & \multicolumn{1}{c}{$\mu(b \rightarrow a)$} \\\hline
		\multirow{2}{*}{U. P} & Boundary & {{4.77}} & \textbf{9.13} & {\textbf{1.05}} & {0.72} \\
		& Non-boundary & {{4.00}} & \textbf{7.00} & {\textbf{0.61}} & {0.39} \\\hline
		\multirow{2}{*}{I.P} & Boundary & {7.84} & \textbf{8.52} & \textbf{4.72} & 1.77 \\
		& Non-boundary & \textbf{7.28} & 5.96 & \textbf{2.64} & 0.54 \\\hline
		\multirow{2}{*}{S} & Boundary & \textbf{2.63} & 2.12 & \textbf{2.85} & 1.39 \\
		& Non-boundary & {2.41} & \textbf{2.44} & \textbf{2.02} & 1.36\\\hline
	\end{tabular}
}
}
\label{tab:mu}
\end{table}
\ifCLASSOPTIONcaptionsoff
  \newpage
\fi


%\bibliographystyle{IEEEtran}
%\FloatBarrier

\vskip 0pt plus -0.9fil
%\vspace{-0.7cm}
\begin{IEEEbiography}[{\includegraphics[width=1in,height=1in,clip,keepaspectratio]{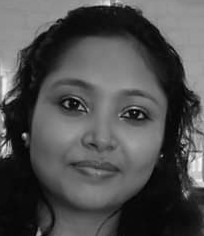}}]{Jayasree Saha} received M.Tech degree from University of Calcutta in 2014. She is currently pursuing the Ph.D degree with the Department of Computer Science and Engineering, Indian Institute of Technology (IIT), Kharagpur, India. Her research interest includes computer vision, remote sensing and deep learning.
\end{IEEEbiography}
\vspace{-15mm}
%\vskip 12pt plus -1.2fil
\begin{IEEEbiography}[{\includegraphics[width=1in,height=1in,clip,keepaspectratio]{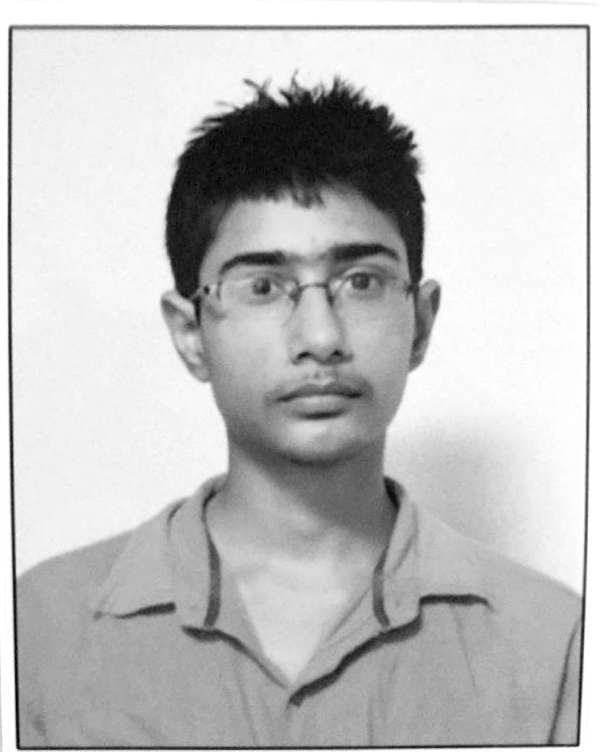}}]{Yuvraj Khanna}
is currently pursuing the Dual degree with the Department of Electronics and Electrical communications and Engineering, Indian Institute of Technology (IIT), Kharagpur, India. His research interest includes remote sensing and deep learning
\end{IEEEbiography}
%\vskip 0pt plus -0.9fil
%\newpage
%\vspace{-9mm}
\begin{IEEEbiography}[{\includegraphics[width=1in,height=1in,clip,keepaspectratio]{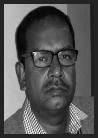}}]{Jayanta Mukherjee}
received his B.Tech., M.Tech., and Ph.D. degrees in Electronics and Electrical Communication Engineering from the Indian Institute of Technology (IIT), Kharagpur in 1985, 1987, and 1990, respectively. He is currently a Professor with the Department of Computer Science and Engineering, Indian Institute of Technology (IIT), Kharagpur, India.  His research interests are in image processing, pattern recognition, computer graphics, multimedia systems and medical informatics. He has published about 250 research papers in journals and conference proceedings in these areas. He received the Young Scientist Award from the Indian National Science Academy in 1992. Dr. Mukherjee is a Senior Member of the IEEE. He is a fellow of the Indian National Academy of Engineering (INAE).
\end{IEEEbiography}


%\vfill


\end{document}